\documentclass{article}

\usepackage{amsmath, amssymb, amsthm}
\usepackage{physics}

\newcommand{\R}{\mathbb{R}} 

\newcommand{\Set}{\mathcal{S}}

\newcommand{\Lamb}{\boldsymbol{\lambda}}

\newcommand{\D}{\mathbf{D}}
\newcommand{\I}{\mathbf{I}}
\newcommand{\path}{\mathbf{p}}
\newcommand{\Path}{\mathbf{P}}
\newcommand{\M}{\mathbf{M}}
\newcommand{\W}{\mathbf{W}}
\newcommand{\T}{\mathbf{T}}
\newcommand{\m}{\mathbf{m}}

\newcommand{\vi}{\mathbf{v}}
\newcommand{\x}{\mathbf{x}}
\newcommand{\y}{\mathbf{y}}
\newcommand{\z}{\mathbf{z}}
\newcommand{\bias}{\mathbf{b}}
\newcommand{\eps}{\boldsymbol{\epsilon}}
\newcommand{\low}{\mathbf{l}}
\newcommand{\up}{\mathbf{u}}

\usepackage{microtype}
\usepackage{graphicx}
\usepackage{subfigure}
\usepackage{booktabs} 
\usepackage{xcolor}

\usepackage{algorithmic}
\usepackage{amsmath}
\usepackage{multicol}
\usepackage{makecell}

\usepackage{tabu}
\usepackage{multirow}

\usepackage{hyperref}

\usepackage[accepted]{icml2019}

\icmltitlerunning{On Certifying Non-uniform Bounds against Adversarial Attacks}

\begin{document}

\twocolumn[
\icmltitle{On Certifying Non-uniform Bounds against Adversarial Attacks}



\icmlsetsymbol{equal}{*}

\begin{icmlauthorlist}
\icmlauthor{Chen Liu}{EPFL}
\icmlauthor{Ryota Tomioka}{MSR}
\icmlauthor{Volkan Cevher}{EPFL}
\end{icmlauthorlist}

\icmlaffiliation{EPFL}{EPFL, Lausanne, Switzerland}
\icmlaffiliation{MSR}{Microsoft Research, Cambridge, UK}

\icmlcorrespondingauthor{Chen Liu}{chen.liu@epfl.ch}

\icmlkeywords{Optimization, Deep Learning}

\vskip 0.3in
]



\printAffiliationsAndNotice{}  

\begin{abstract}
This work studies the robustness certification problem of neural network models,
which aims to find certified adversary-free regions as large as possible around data points.
In contrast to the existing approaches that seek regions bounded uniformly along all input features,
we consider non-uniform bounds and use it to study the decision boundary of neural network models.
We formulate our target as an optimization problem with nonlinear constraints.
Then, a framework applicable for general feedforward neural networks is proposed to bound the output logits
so that the relaxed problem can be solved by the augmented Lagrangian method.
Our experiments show the non-uniform bounds have larger volumes than uniform ones and the geometric similarity of the non-uniform bounds gives a quantitative, data-agnostic metric of input features' robustness.
Further, compared with normal models, the robust models have even larger non-uniform bounds and better interpretability.
\end{abstract}

\section{Introduction}

Although deep neural networks have achieved great success and state-of-the-art performances in many tasks, they are vulnerable to some adversarial attacks of the input data~\cite{szegedy2013intriguing, goodfellow2015explaining, moosavi2017universal}.
This issue can be contextualized as a game between the attackers who try to adversarially manipulate the input data and the defenders who try to obtain the robust model parameters.

Numerous methods have been proposed to attack or defend deep neural network models.
Popular attack methods include fast gradient method (FGM) \cite{goodfellow2015explaining}, iterative fast gradient method (IFGM) \cite{kurakin2016adversarial}, projected gradient descent (PGD) \cite{madry2017towards} and CW attack \cite{carlini2017towards}.
Most attack methods search for the adversarial example by utilizing the gradient of loss objective w.r.t. the input data.
On the defenders' side, adversarial training \cite{szegedy2013intriguing}, which augments the training data with adversarial examples, is a simple and popular method.
It achieves the best empirical performance over other recent methods \cite{buckman2018thermometer, ma2018characterizing, guo2017countering, dhillon2018stochastic, xie2017mitigating, song2017pixeldefend, samangouei2018defense} under adversarial settings in \citet{athalye2018obfuscated}.

However, recent work \cite{athalye2018obfuscated} shows that these \textit{uncertified} defense methods might fail when a stronger attack is applied.
Therefore, finding a \textit{provable} defense algorithm and \textit{certifying} the level of robustness of an input data point have become active research topics.
We need to guarantee that the models output consistent results in the worst cases under some conditions.

Following this line of works, we focus on the certification problem in this paper:

\textit{
Given the label set $\mathcal{C}$, a classification model $f: \R^{n} \to \mathcal{C}$ and an input data point $\x \in \R^{n}$, we would like to find the largest neighborhood $\Set$ around $\x$ such that $f(\x) = f(\x')\ \forall \x' \in \Set$.
}

Many methods have been proposed for certifying a region bounded around a data point.
Early studies use Satisfiability Modulo Theories (SMT) solvers \cite{huang2017safety} or integer programming approaches \cite{lomuscio2017approach}.
Despite some recent progress, these combinatorial methods still tend to suffer from superpolynomial time complexity \cite{katz2017reluplex, wang2018efficient} or requires solving a semi-definite programming problem \cite{raghunathan2018certified, raghunathan2018semidefinite}.

More recently, \citet{kolter2017provable} and \citet{wong2018scaling} use a convex polytope relaxation to construct the bounds of the model's output logits.
The gap between the logits of true and false labels is then minimized by primal-dual methods and can be trained by a dual network.
\citet{zhang2018efficient} generalizes the method of \citet{kolter2017provable} to non-ReLU networks.
\citet{singh2018fast} further proposes a method of higher speed and precision based on the abstract interpretation of neural networks.
These methods generally have \textit{quadratic} complexity for strict bound.

All the studies mentioned above concentrate on certifying a \textit{uniform} bound around a data point against adversarial perturbations.
That is say, the certified regions $\Set$ of all these algorithms are \textit{uniform} across all input features.
For example, when the input has two dimensions, the certified region is a square under $l_\infty$ norm and a perfect round circle under $l_2$ norm. 

In this paper, we explore the possibility of finding a \textit{non-uniform} bounded data neighborhood without adversaries.
In two-dimensional case above, the certified region would be a rectangle under $l_\infty$ norm and an elliptical under $l_2$ norm.
Such different level of robustness among input features has already found interest in the literature, such as \citet{tsipras2018there}.
Indeed, robust features should have larger perturbation tolerance than non-robust ones.
Under \textit{non-uniform} settings, we can have larger certified regions and a quantitative robustness metric for different input features.

Perhaps more importantly, we can use \textit{non-uniform} bound as a tool to study the decision boundaries of different models.
This is fundamental regarding understanding the robustness property of neural network models.
Our work is an important step towards that.

We summarize the contributions of our work below:

\begin{itemize}
    \item We provide a framwork to estimate the bounds of output logits in general feedforward neural network given non-uniform adversarial budget $\eps$.
    Any existing bounding estimation algorithm can fit this framework as long as its bound is diferentiable w.r.t. $\eps$
    \item In order to find the certified non-uniform bounds of the largest volumes, we formulate the goal as a constrained optimization problem, make relaxations and solve the relaxed problem by the augmented Langrangian method.
    \item Our method can find non-uniform bounds of larger volumes than uniform ones, revealing at least three benefits of robust models: certified non-uniform bounds of larger volumes, better interpretability and higher geometric similarity.
\end{itemize}

We formalize our non-uniform bound certification problem in Section \ref{sec:problem} and propose solution algorithms in Section \ref{sec:alg}.
We provide experimental evidence in Section \ref{sec:exp}, followed up by extensions and future works in Section \ref{sec:discussion}. Conclusions can be found in Section \ref{sec:conclusion}.

\section{Problem Formulation} \label{sec:problem}

We start with a $N$-layer fully connected neural network, parameterized by $\{\W^{(i)}, \bias^{(i)}\}_{i = 1}^{N - 1}$:

\begin{equation}
\begin{aligned}
\z^{(i + 1)} &= \W^{(i)} \hat{\z}^{(i)} + \bias^{(i)} &i &= 1, 2, ..., N - 1 \\
\hat{\z}^{(i)} &= \sigma(\z^{(i)}) &i &= 2, 3, ..., N - 1 \\
\end{aligned}\label{eq:nn}
\end{equation}

where $\{\z^{(i)}, \hat{\z}^{(i)}\}_{i = 1}^{N - 1}$ are pre- and post- activation values in each layer.
The input data and the output logits of the neural network are $\hat{\z}^{(1)}$ and $\z^{(N)}$ respectively.
$\sigma$ is a nonlinear function which can be ReLU, sigmoid, tanh, etc.
We use $n_1, n_2, ..., n_N$ to denote the number of neurons in each layer, so $\W^{(i)} \in \R^{n_{i + 1} \times n_i}$ and $\bias^{(i)} \in \R^{n_{i + 1}}$

In this work, the adversarial budget for an input data point $\x \in \R^{n_1}$ is represented by a non-uniform bounded region $\Set_{\eps}^{(p)}(\x)$, parameterized by $\eps \in \R^{n_1}$.
We define set $\Set_{\eps}^{(p)}(\x)$ as $\{\hat{\z}^{(1)} = \x + \eps \odot \mathbf{v} | \|\mathbf{v}\|_p \leq 1\}$ based on $l_p$ norm.
In most parts of this work, we focus on the $l_\infty$ case and discuss the potential extension to other norms later.
For simplicity, we use $\Set_{\eps}(\x)$ to represent $\Set_{\eps}^{(\infty)}(\x)$.

Now, we would like to find the certified region $\Set_{\eps}(\x)$ of the largest volume, measured by $\Pi_{j = 0}^{n_1 - 1} {\epsilon}_j$, in which the model outputs consistent label.
Formally, for a data point labeled as the category $c \in \{0, 1, ..., n_N - 1\}$, the problem we focus on is formulated below:

\begin{equation}
\begin{aligned}
 \min_{\eps} \quad &\left\{ -\sum_{j = 0}^{n_1 - 1} \log \epsilon_j \right\} \\
\hat{\z}^{(1)} &\in \Set_{\eps}(\x) \\
\z^{(i + 1)} &= \W^{(i)} \hat{\z}^{(i)} + \bias^{(i)}  & &i = 1, 2, ..., N - 1 \\
\hat{\z}^{(i)} &= \sigma(\z^{(i)})  & &i = 2, 3, ..., N - 1 \\
z^{(N)}_c - z^{(N)}_j &\geq \delta  &j = 0, 1, &..., n_N - 1; j \neq c
\end{aligned}\label{eq:problem}
\end{equation}

As a common practice, we minimize the negative logarithm of $\Pi_{j = 0}^{n_1 - 1} {\epsilon}_j$ because it is convex and more stable numerically.
$\delta$ is a small positive constant to make sure the logits of the true label is strictly higher than others.
We can see that if we constrain $\eps$ by a scalar $\gamma$: $\eps = \gamma \mathbf{1}$, the problem can be reduced to uniform bound certification problem, the one solved by \citet{kolter2017provable, wong2018scaling, raghunathan2018certified, zhang2018efficient, singh2018fast}.

Before going into the details of the algorithm, we also introduce the notation used in the sequel.
We use $\low^{(i)}$ and $\up^{(i)}$ to represent the lower and upper bound of pre-activation values in the $i$-th layer respectively i.e. $\low^{(i)} \leq \z^{(i)} \leq \up^{(i)}$.
For a tensor $\T$, $\T_{-}$ means all its negative elements i.e. $\T_{-} = \min(\T, 0)$. Similarity, we can define $\T_{+}$.
$\odot$ is elementwise product operation between two tensors.
Bracketed superscripts are used to index tensors while subscripts mean the elements in a tensor.
Scalars are broadcast when it is added or subtracted from tensors.
The equality and inequality relations in this paper are all elementwise.


\section{Algorithm} \label{sec:alg}

In this section, following the work of \citet{kolter2017provable, singh2018fast, zhang2018efficient}, we derive a
linear approximation to bound the nonlinear activation function $\sigma$. This allows us to relax problem (\ref{eq:problem}) into
an optimization problem with bounds parameterized as a (nonlinear) function of ${\eps}$. Then we show that we can compute the gradient 
of this bound with respect to ${\eps}$ efficiently and solve the relaxed problem using the augmented Lagrangian method.

\subsection{Linear Approximation of Activation Functions} \label{subsection:nonlinear_approx}


For an activation function $\sigma(x)$, which is nonlinear and monotonic, and $x$ bounded by $l \leq x \leq u$, we can linearize it by two linear functions with the same slope: $kx + m_1 \leq \sigma(x) \leq kx + m_2$. 
$k$, $m_1$ and $m_2$ all depend on the bounds $l$ and $u$.
They are chosen in a way such that the gap $m_2 - m_1$ between bias terms shall be as small as possible.

For example, if $\sigma$ is ReLU function: $\sigma(x) = \max(0, x)$, we have:

\begin{equation}
\begin{aligned}
k = \left\{
\begin{aligned}
&0 & l \leq u \leq 0 \\
&\frac{u}{u - l} & l < 0 < u \\
&1 & 0 \leq l \leq u
\end{aligned}
\right.&,
m_2 = \left\{
\begin{aligned}
&0 & l \leq u \leq 0 \\
&- \frac{ul}{u - l} & l < 0 < u \\
&0 & 0 \leq l \leq u
\end{aligned}
\right. \\
m_1 &= 0
\end{aligned}
\end{equation}

Here, we find $m_1 = m_2$ when $l \leq u \leq 0$ or $0 \leq l \leq u$.
Therefore, the linear approximation error arises for ReLU only when $l < 0 < u$.

For a general activation function $\sigma(x)$, we can consider functions $\tilde{m}_1(k) \leq \sigma(x) - kx$ and $\tilde{m}_2 \geq \sigma(x) - kx$ when $x \in [l, u]$.
Then $k$ is chosen by minimizing the margin $\tilde{m}_2 - \tilde{m}_1$.
\citet{zhang2018efficient} extends \citet{kolter2017provable}'s methods in a similar way but they use numerical methods to obtain $k$.
Here $k$ needs to have analytical form, because we need the gradients of bounds w.r.t. $\eps$.

Consider a vector $\x$ bounded by $\low \leq \x \leq \up$, we can bound $\sigma(\x)$ by $\D \x + \m_1 \leq \sigma(\x) \leq \D \x + \m_2$ where $\D$ is a diagonal matrix and $\m_1$, $\m_2$ are bias vectors.
Equivalently, we can say $\forall \x: \low \leq \x \leq \up$, $\exists \m: \m_1 \leq \m \leq \m_2$ such that $\sigma(\x) = \D \x + \m$.

For a fully connected neural network defined in equation (\ref{eq:nn}), we linearize $\hat{\z}^{(i)}$ by $\hat{\z}^{(i)} = \sigma(\z^{(i)}) = \D^{(i)}\z^{(i)} + \m^{(i)}$ under constraint $\m^{(i)}_1 \leq \m^{(i)} \leq \m^{(i)}_2$ given the bound $\low^{(i)} \leq \z^{(i)} \leq \up^{(i)}$.
For the input layer, the matrix $\D^{(1)}$ is identity and the bias term is bounded by $\eps$: $-\eps \leq \m^{(1)} \leq \eps$.
If we unfold the linear approximations, the output of each layer $\z^{(i)}$ can be expressed in the following way:

\begin{equation}
\small
\begin{aligned}
\z^{(i)} &= \W^{(i-1)}(\sigma(\W^{(i-2)}(... \\ &(\W^{(1)}(\x + \m^{(1)}) + \bias^{(1)})...) + \bias^{i -2})) + \bias^{(i-1)} \\
&=  \W^{(i - 1)}(\D^{(i - 1)}(\W^{(i - 2)}(... \\ &(\W^{(1)}(\x + \m^{(1)}) + \bias^{(1)})...) + \bias^{(i - 2)}) + \m^{(i - 1)}) + \bias^{(i - 1)} \\
&= \left(\Pi_{j = 2}^{i - 1}\W^{(j)}\D^{(j)}\right)\W^{(1)}\x + \sum_{h = 1}^{i - 1}\left(\Pi_{j = h + 1}^{i - 1}\W^{(j)}\D^{(j)}\right)\bias^{(h)} \\
&+ \sum_{h = 1}^{i - 1}\left(\Pi_{j = h + 1}^{i - 1}\W^{(j)}\D^{(j)}\right)\W^{(h)}\m^{(h)}
\end{aligned} \label{eq:zi}
\end{equation}

\subsection{Lower and Upper Bound Estimation} \label{subsection:bound_est}

According to equation (\ref{eq:zi}), for the output layer or any intermediate layer $\z^{(i)}$ of a model, the only variables on the right hand side are $\{\m^{(h)}\}_{h = 1}^{i - 1}$.
The bound of $\z^{(i)}$ can be obtained immediately from the bounds of $\{\m^{(h)}\}_{h = 1}^{i - 1}$.
Once we have the bound of $\z^{(i)}$, we can then obtain $\D^{(i)}$, $\m^{(i)}_1$ and $\m^{(i)}_2$ to bound $\z^{(i + 1)}$.
Such process can be done iteratively until we obtain the lower bound $\low^{(N)}$ and upper bound $\up^{(N)}$ of the output logits $\z^{(N)}$.
The computational complexity of FLOP operations in this algorithm is $O(N^2 n^3)$ where $n$ is the maximum number of neurons in a hidden layer.
We call this algorithm \textit{quadratic algorithm} and provide pseudo code in \textbf{Algorithm} \ref{alg:complex_bound_est} in Appendix \ref{sec:app_alg}.
This algorithm is the same as CROWN in \citet{zhang2018efficient} except that we use the same slope for upper and lower bound of the nonlinearity.
\textbf{Algorithm} \ref{alg:complex_bound_est} only maintains one list of matrices i.e. $\M^{(j)}$ in each iteration while CROWN needs 4 such matrices and thus consumes much more memory.

Theoretically, any convex hull of the nonlinear function $\sigma$ in $[l, u]$ leads to a valid bound of the output logits \cite{salman2019convex}, therefore, the smallest convex hull corresponds to the tightest bound.
However, in practice we can hardly find an analytical form of the smallest convex hull for arbitrary $\sigma$.
It is a trade-off between the tightness and the complexity.
In this work, we use two parallel line to bound the nonlinear function within a given range, so our convex hull here is a parallelogram.

On the other hand, we can obtain a naive layerwise bound by iteratively calculating the bound of a layer based on the bound of the immediate previous layer.
Compared with \textit{quadratic algorithm}, the complexity of FLOP operations in this algorithm is $O(N n^2)$ \footnote{In \textit{simple algorithm}, there are only matrix-vector multiplications, whose complexity are $O(n^2)$ each.} and we call it \textit{simple algorithm}
The pseudo code is in \textbf{Algorithm} \ref{alg:naive_bound_est} in Appendix \ref{sec:app_alg}.
Because of its efficiency, \textit{Simple algorithm} has been incorporated into the algorithms in \citet{gowal2018effectiveness} to train robust models.

Although \citet{kolter2017provable} has empirically showed that \textit{quadratic algorithm} typically obtains better bounds than \textit{simple algorithm}, we find that they are actually complementary.
On one hand, \textit{quadratic algorithm} linearizes the activation functions and the gap between upper and lower bounds come from the flexibility of $\{\m^{(h)}\}_{h = 1}^{i - 1}$, which grows moderately with the number of layers.
\textit{Simple algorithm} calculates the bounds layerwisely and the error can propagate much faster.
However, \textit{simple algorithm} calculates the activation function exactly without any approximation.
Therefore, neither algorithm is \textit{guaranteed} to be better than the other.
Roughly speaking, \textit{simple algorithm} works better for networks of few layers (`shallow' networks) or networks containing layers of very few neurons (`thin' networks), while \textit{quadratic algorithm} is better in other cases.
Detailed discussion and examples for comparison are deferred in Appendix \ref{sec:app_discussion}.

In this work, we incorporate \textit{simple algorithm} into \textit{quadratic algorithm} for a better bound in each layer.
The pseudo code is given as \textbf{Algorithm} \ref{alg:bound_est} below.
The maximum and minimum operators in line 15 and 16 are applied elementwisely.

\begin{algorithm}
\small
\begin{algorithmic}[1]
\STATE Input: Parameters $\{\W^{(i)}, \bias^{(i)}\}_{i = 1}^{N - 1}$, perturbation set $\Set_{\eps}(\x)$.
\STATE $\low^{(2)} = \W^{(1)}\x - \W^{(1)}_{+}\eps + \W^{(1)}_{-}\eps + \bias^{(1)}$
\STATE $\up^{(2)} = \W^{(1)}\x - \W^{(1)}_{-}\eps + \W^{(1)}_{+}\eps + \bias^{(1)}$
\STATE $\M^{(1)} = \W^{(1)}$
\STATE $\phi^{(2)} = \W^{(1)}\x + \bias^{(1)}$
\FOR {i = $2, ..., N -1$}
    \STATE Calculate $\D^{(i)}$, $\m_1^{(i)}$, $\m_2^{(i)}$ based on $\low^{(i)}$ and $\up^{(i)}$
    \STATE $\low^{(i + 1)}_{simp} = \W^{(i)}_{+}\sigma(\low^{(i)}) + \W^{(i)}_{-}\sigma(\up^{(i)})$
    \STATE $\up^{(i + 1)}_{simp} = \W^{(i)}_{-}\sigma(\low^{(i)}) + \W^{(i)}_{+}\sigma(\up^{(i)})$
    \STATE $\M^{(j)} = \W^{(i)}\D^{(i)}\M^{(j)}$ for $j = 1, ..., i - 1$
    \STATE $\M^{(i)} = \W^{(i)}$
    \STATE $\phi^{(i + 1)} = \W^{(i)}\D^{(i)}\phi^{(i)} + \bias^{(i)}$
    \STATE $\low^{(i + 1)}_{quad} = \phi^{(i + 1)} + \sum_{j = 1}^{i} \left( \M^{(j)}_{-}\m_2^{(j)} + \M^{(j)}_{+}\m_1^{(j)} \right)$
    \STATE $\up^{(i + 1)}_{quad} = \phi^{(i + 1)} + \sum_{j = 1}^{i} \left( \M^{(j)}_{-}\m_1^{(j)} + \M^{(j)}_{+}\m_2^{(j)} \right)$
    \STATE \label{line:tighten1} $\low^{(i + 1)} = \max(\low^{(i + 1)}_{simp}, \low^{(i + 1)}_{quad})$
    \STATE \label{line:tighten2} $\up^{(i + 1)} = \min(\up^{(i + 1)}_{simp}, \up^{(i + 1)}_{quad})$
\ENDFOR
\STATE Output: Bounds $\{\low^{(i)}, \up^{(i)}\}_{i = 2}^{N}$
\end{algorithmic}
\caption{Bound Estimation} \label{alg:bound_est}
\end{algorithm}




\subsection{Gradient of Perturbation Budget $\eps$}

\textbf{Algorithm} \ref{alg:bound_est} provides the algorithm to estimate the bound of output logits for any input perturbation $\Set_{\eps}(\x)$.
In uniform bound certification problem, we constrain $\eps = \gamma \mathbf{1}$ and have one dimensional variable $\gamma \in \R$ to optimize.
In this case, the optimality can be found by binary search or line search \cite{zhang2018efficient}.
However, in non-uniform bound certification problem, we have $n_1$ dimensional variable $\eps$ to optimize.
It becomes necessary to estimate a good direction to update $\eps$.
This is why we consider gradient methods, and fortunately the (sub)gradient $\frac{\partial \low^{(N)}}{\partial \eps}$ and $\frac{\partial \up^{(N)}}{\partial \eps}$ can obtained according to \textbf{Algorithm} \ref{alg:bound_est}.

Based on the linear approximation as discussed in Section \ref{subsection:nonlinear_approx}, standard back-propagation can be applied to calculate the (sub)gradients of the final bounds w.r.t. $\eps$.
Alternatively, similar to \textbf{Algorithm} \ref{alg:bound_est}, the gradient can be calculated recursively using chain rule.
Unlike back-propagation, this method does not need to wait for the termination of \textbf{Algorithm} \ref{alg:bound_est} and can be calculated on the fly, which is beneficial in the distributed settings.
The pseudo code is provided in Appendix \ref{sec:app_alg}.

\subsection{Optimization by the Augmented Lagrangian Method}

Now, we have done all the preparation to solve the relaxation of problem (\ref{eq:problem}).
By bound estimation, we rewrite the problem below:

\begin{equation}
\begin{aligned}
&\min_{\eps, \y \geq 0} \left\{ - \sum_{j = 0}^{n_1 - 1} \log \epsilon_j \right\}\\
& s.t.\ l^{(N)}_c - \up^{(N)}_{j \neq c} - \delta = \y
\end{aligned}\label{eq:problem_rewrite}
\end{equation}

Here $\low^{(N)}$ and $\up^{(N)}$ are functions of $\eps$ given by \textbf{Algorithm} \ref{alg:bound_est}.
$\up^{(N)}_{j \neq c} \in \R^{n_N - 1}$ is the concatenation of all output logits except true label $c$.
$\y (\geq 0) \in \R^{n_N - 1}$ is a slack variable that ensures the nonnegativity of the term on the left hand side.
For simplicity, we define $\vi := l^{(N)}_c - \up^{(N)}_{j \neq c} - \delta$ as a function of $\eps$.

Note that we have replaced constraint $\z^{(N)}_c - \z^{(N)}_j \geq \delta$ in problem (\ref{eq:problem}) by a stronger version
$l^{(N)}_c - \up^{(N)}_{j \neq c} \geq \delta$.
Therefore, the optimality of problem (\ref{eq:problem_rewrite}) provides the upper bound of the original minimization problem (\ref{eq:problem}).

We can further rewrite the problem (\ref{eq:problem_rewrite}) into a min-max problem using augmented Lagragian method~\cite{hestenes1969multiplier, powell1969method} by introducing the dual variable $\Lamb \in \R^{n_N - 1}$ and the coefficient $\rho \in \R^{+}$.
The dual problem to solve is below:

\begin{equation}
\begin{aligned}
\max_{\Lamb} \min_{\eps, \y \geq 0} - \left( \sum_{j = 0}^{n_1 - 1} \log \epsilon_j \right) &+ \left\langle \Lamb, \vi - \y \right\rangle + \frac{\rho}{2} \|\vi - \y\|_2^2
\end{aligned} \label{eq:al}
\end{equation}

The inner minimization problem is a quadratic form of $\y$, so the optimal $\y$ has the analytical solution: $\y = \max(0, \vi + \frac{1}{\rho}\Lamb)$.
Plug the solution in the problem and we can optimize $\eps$ by gradient descent.
The pseudo code is in \textbf{Algorithm} \ref{alg:al}.

\begin{algorithm}
\small
\begin{algorithmic}[1]
\STATE Input: Parameters $\{\W^{(i)}, \bias^{(i)}\}_{i = 1}^{N - 1}$, original bounds $\eps_0$, iterations $M$, augmented coefficient $\{\rho^{(i)}\}_{i = 1}^{M}$, decaying factor $\eta$.
\STATE $\eps = \eps_0$, $\Lamb = 0$
\FOR {$i = 1, 2, ..., M$}
    \STATE \label{line:al_min} Update $\eps$ by minimizing (\ref{eq:al}) with optimal $\y$ plugged in.
    \STATE $\Lamb = \Lamb + \rho^{(i)} (\vi - \y)$
\ENDFOR
\WHILE {$\vi \geq 0$ is not satisfied} \label{line:while}
    \STATE $\eps = \eta \eps$
\ENDWHILE
\STATE Output: $\eps$
\end{algorithmic}
\caption{Optimization for $\eps$} \label{alg:al}
\end{algorithm}

Similar to penalty method, the coefficient $\{\rho^{(i)}\}_{i = 1}^{M}$ in \textbf{Algorithm} \ref{alg:al} is a non-decreasing sequence to enforce the constraint.
However, the Lagrange multiplier term makes it unnecessary to increase $\rho^{(i)}$ to $+ \infty$.
Actually, $\rho^{(i)}$ can stay much smaller here to solve the problem, which avoids numerical instability caused by ill-conditioning.

The minimization in line 4 is solved by gradient methods.
In practice, gradient explosion might happen when $\eps$ is small or $\rho^{(i)}$ is big.
To avoid overshooting, we apply gradient rescaling to constrain the $l_2$ norm of the gradient.
The term $\log\epsilon_j$ implicitly constrains $\epsilon_j$ to be positive, so we reparametrize $\epsilon_j = \zeta_j^2$ and optimize vector $\boldsymbol{\zeta}$.

The last while-loop in line 7 is to ensure the output $\eps$ meets the hard constraints.
The decaying factor $\eta$ is close to $1$ and is set $0.99$ in practice.
When $\rho^{(i)}$ is large, the while-loop would break after very few iterations.

\section{Experiments} \label{sec:exp}

In this Section, we compare our certified non-uniform bounds with uniform bounds.
We also use our algorithm as a tool to explore the decision boundaries of different models.
All the experiments here are implemented in the framework of PyTorch and can be finished within several hours on a single NVIDIA Tesla GPU machine.

Because any algorithm of estimating the output logits can be incorporated into our framework of computing non-uniform bounds, our main focus in this section is the comparison between uniform and non-uniform bounds based on the same estimation algorithm of output logits.

\subsection{Synthetic Data}

We first validate our algorithm using 2-D synthetic data so that we can visualize the certified bounds.

We generate 10 random 2D data points in the space of $[-1, 1]^2$ labeled $\{0, 1, ..., 9\}$ as seeds.
Another $10000$ random points in $[-1, 1]^2$ are then generated and assigned the same label as the closest seeds.
$90\%$ of the data points are in the training set and the rest are reserved for testing.

The model here is a ReLU fully-connected neural network with two hidden layers of 10 neurons.
Since the boundary between different categories are piecewise linear in this case, the model is shown to have enough capacity and achieve an accuracy of more than $99.9\%$ in the test set.

\begin{figure}
\centering
\includegraphics[scale = 0.48]{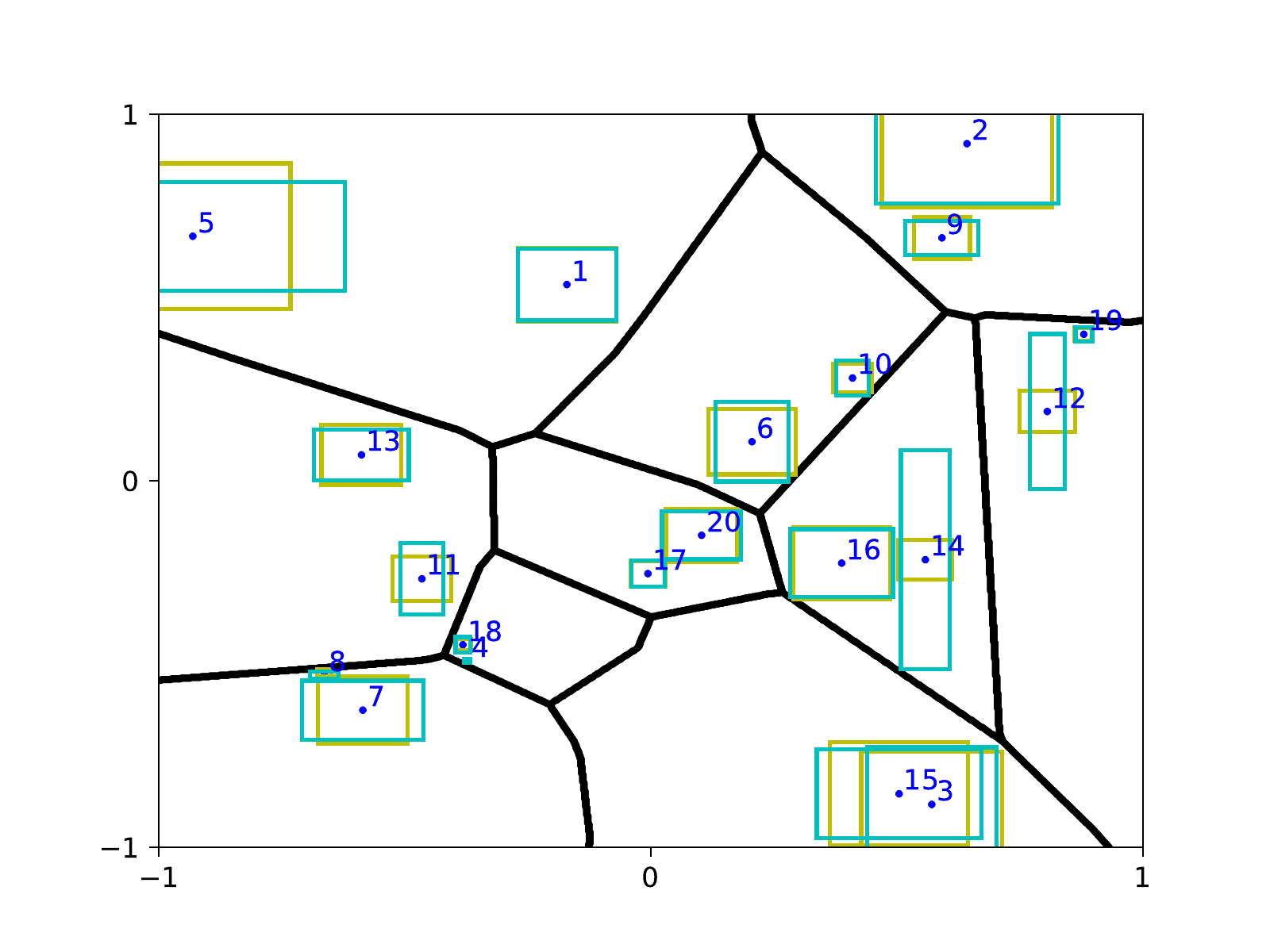}
\caption{A simple example on the synthetic data. 20 points with their certified uniform (yellow) and non-uniform (blue) bounds are shown above. The real decision boundary is shown as black lines.}\label{fig:synthetic_bounds}
\end{figure}

Figure \ref{fig:synthetic_bounds} demonstrates the results of uniform (yellow) and non-uniform (blue) bounds with two random points in each category.
We can clearly see the bounds calculated by our algorithm are reasonably tight and areas covered by non-uniform bounds are larger than those of uniform bounds.
Although the larger volumes do not necessarily mean the bounds are larger for both features, bounds of some feature are extended in compensation for the other.
We can say the features are more robust when their bounds are larger.

In some cases, the non-uniform bounds can be significantly larger than uniform bounds (e.g. point 12, 14 in Figure \ref{fig:synthetic_bounds}).
This typically means considerable difference in robustness between two input features.

\begin{table*}[t]
\centering
\begin{tabular}{|c|c|c|c|c|c|c|}
\hline
Dataset & Architecture & Adversary & Accuracy (\%) & Uniform Bound & Non-uniform Bound & Ratio\\
\hline
\multirow{6}{*}{MNIST} & \multirow{2}{*}{100-100-100} & -  & 99.2 & 0.0295 & 0.0349 & 1.183 \\
\cline{3-7}
& & PGD, $\tau = 0.1$ & 98.1 & 0.0692 & 0.1678 & 2.425 \\
\cline{2-7}
& \multirow{2}{*}{300-300-300} & - & 98.0 & 0.0309 & 0.0350 & 1.133 \\
\cline{3-7}
& & PGD, $\tau = 0.1$ & 98.9 & 0.0507 & 0.1404 & 2.769 \\
\cline{2-7}
& \multirow{2}{*}{500-500-500} & - & 98.5 & 0.0319 & 0.0360 & 1.129 \\
\cline{3-7}
& & PGD, $\tau = 0.1$ & 98.8 & 0.0436 & 0.1167 & 2.677 \\
\hline
\multirow{2}{*}{\makecell{Fashion \\ MNIST}} & \multirow{2}{*}{1024-1024-1024} & - & 90.4 & 0.0134 & 0.0141 & 1.052 \\
\cline{3-7}
& & PGD, $\tau = 0.1$ & 88.4 & 0.0208 & 0.0306 & 1.468 \\
\hline
\multirow{2}{*}{SVHN} & \multirow{2}{*}{1024-1024-1024} & - & 84.3 & 0.0022 & 0.0072 & 3.273 \\
\cline{3-7}
& & PGD, $\tau = 0.1$ & 78.2 & 0.0054 & 0.0281 & 5.204 \\
\hline
\end{tabular}
\caption{Average of uniform and non-uniform bounds in the test sets. The architecture column means the number of neurons in hidden layers. The accuracy column means the value of clean accuracy. The ratio is the values of non-uniform bounds over uniform bounds.} \label{tbl:general_results}
\end{table*}

\subsection{Real Datasets}

In this part, we run our algorithm on real datasets, including MNIST, Fashion-MNIST and SVHN \cite{netzer2011reading}.
All of them are popular benchmarks for image classification and contain tens of thousand images.
MNIST and Fashion-MNIST are $28 \times 28$ gray-scale images, while SVHN are $32 \times 32$ colored images.
Unless specified, all pixel values of images are normalized in the range of $[-1, 1]$.

Besides a metric of feature robustness, the non-uniform bounds also can be used as a tool to explore the decision boundaries of models.
In the following subsections, we investigate the difference between the decision boundaries of robust and non-robust models.

\subsubsection{Robustness and Volume of Bounds} \label{sec:vol_bounds}

As Table \ref{tbl:general_results} shows, we train different models for different datasets in different ways.
To obtain robust models, we do adversarial training based on PGD \cite{madry2017towards} attacks.
To the best of our knowledge, this is the way to obtain the most robust model empirically studied in \citet{athalye2018obfuscated}.
We set the perturbation budget $\tau$ of PGD to be $0.1$ and search for adversarial examples for 20 iterations.
We call models adversarially trained by PGD \textit{robust models} to distinguish from \textit{normal models}.

We report results based on data in the test sets here.
To make the results of non-uniform bound comparable with uniform bound, we take the geometric average values $\left(\Pi_{j = 0}^{n_1 -1} \epsilon_j\right)^{\frac{1}{n_1}}$.
Table \ref{tbl:general_results} shows the average bounds in different settings, it is clear that non-uniform bounds consistently certify areas of larger volumes.

\begin{figure}[h]
\includegraphics[scale = 0.55]{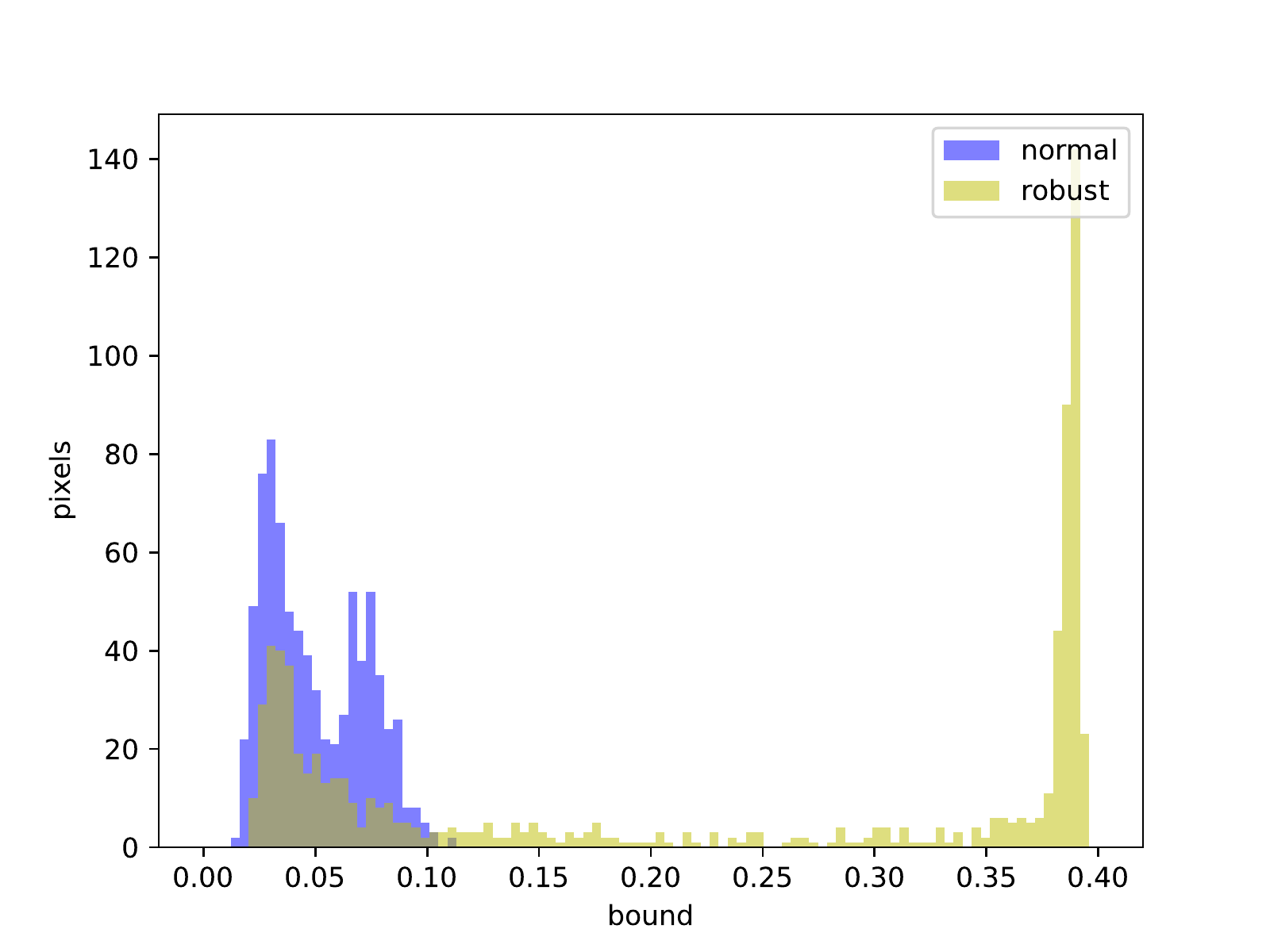}
\caption{The distributions of bounds per feature for normal and robust models in a randomly picked image.
Both models have three hidden layers of 300 neurons.
The uniform bounds for normal and robust models are 0.0402 and 0.0568 respectively,
while the corresponding geometric average values of non-uniform bounds are 0.0450 and 0.1462 respectively.} \label{fig:hist_bounds}
\end{figure}

We notice that the ratio of non-uniform bound over uniform bound is significantly larger in the cases of robust models.
Figure \ref{fig:hist_bounds} shows the histogram of bound per feature for normal and robust models on a randomly picked image in MNIST.
Compared with the normal model, the bounds of some features for the robust model can be as large as $0.4$, much larger than the value of $\tau$.
This observation means the decision boundary of the robust model is almost aligned in some dimensions corresponding to some features,
which makes it possible for our algorithm to extend the bounds of those features without sacrificing the bounds of the others much.
It also implies robust models tend to drop irrelevant features and rely on fewer features when making predictions.
We put more results from other MNIST models as well as models on Fashion-MNIST and SVHN dataset to support our claim in Appendix \ref{sec:app_exp_general}.

\subsubsection{Robustness and Model Interpretability} \label{sec:interpretability}

Given an image and a neural network model, our algorithm can obtain a non-uniform bound parameterized by $\eps \in \mathbb{R}^{n_1}$.
For high dimensional images, we can not plot the rectangular bounds.
However, we can visualize $\eps$ just like images.
We call them \textit{bounding maps} and use the same rescaling factor \footnote{All bounding map figures in this paper are plot based on $\mathbf{1} - 5\eps$, so darker pixels in bounding maps mean larger bounds.} to visualize them in this paper.

\begin{figure}
\centering
\subfigure{\includegraphics[scale = 0.18]{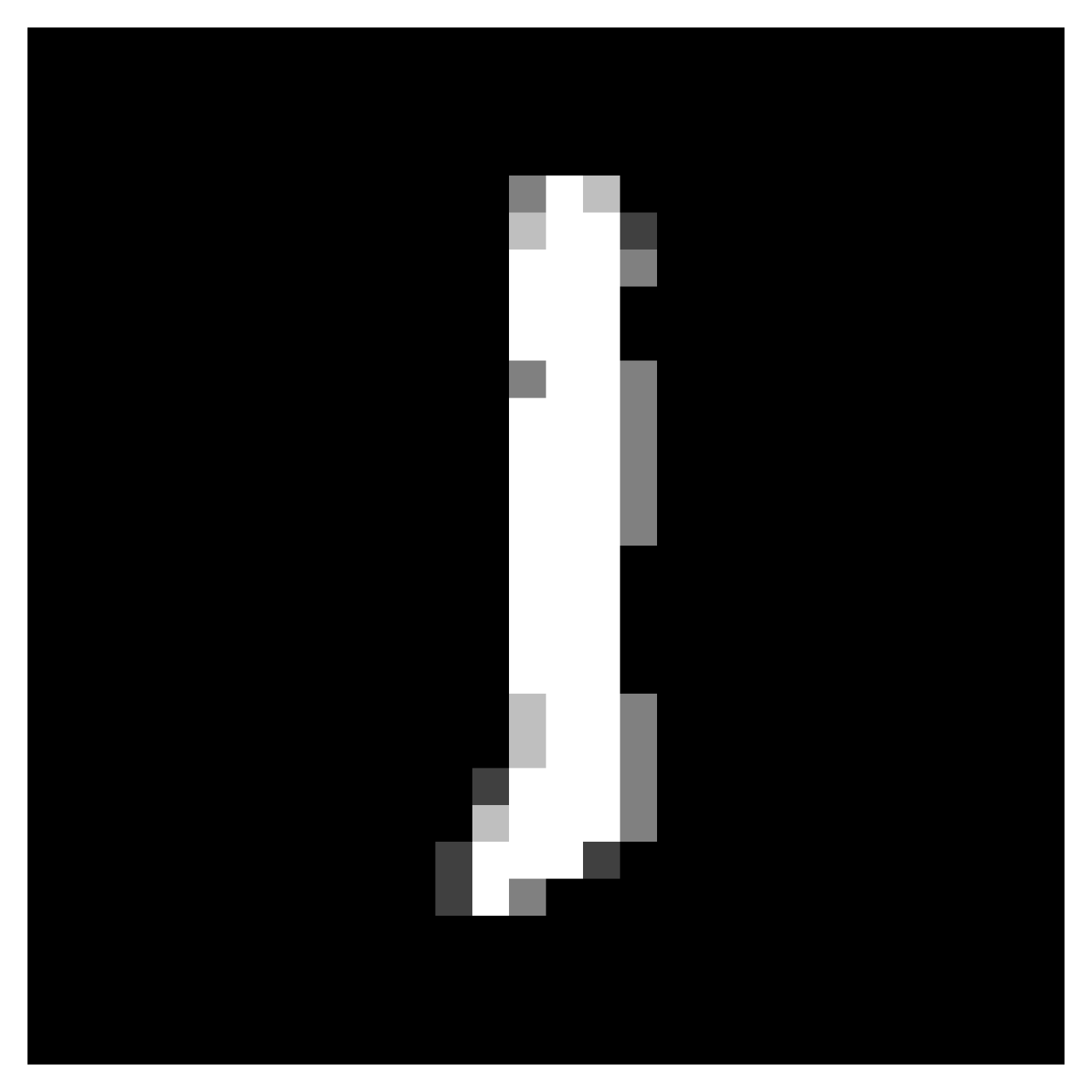}}
\subfigure{\includegraphics[scale = 0.18]{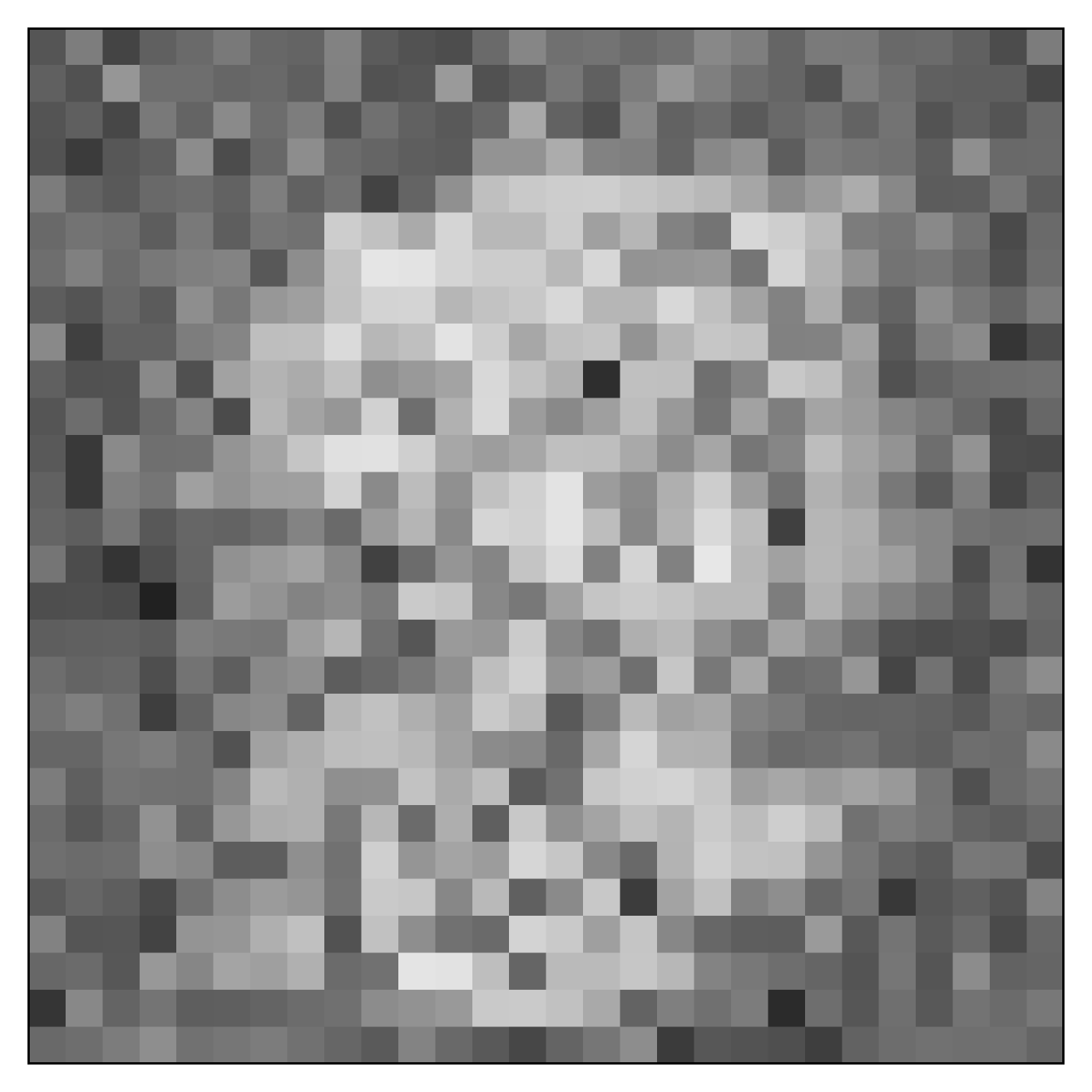}}
\subfigure{\includegraphics[scale = 0.18]{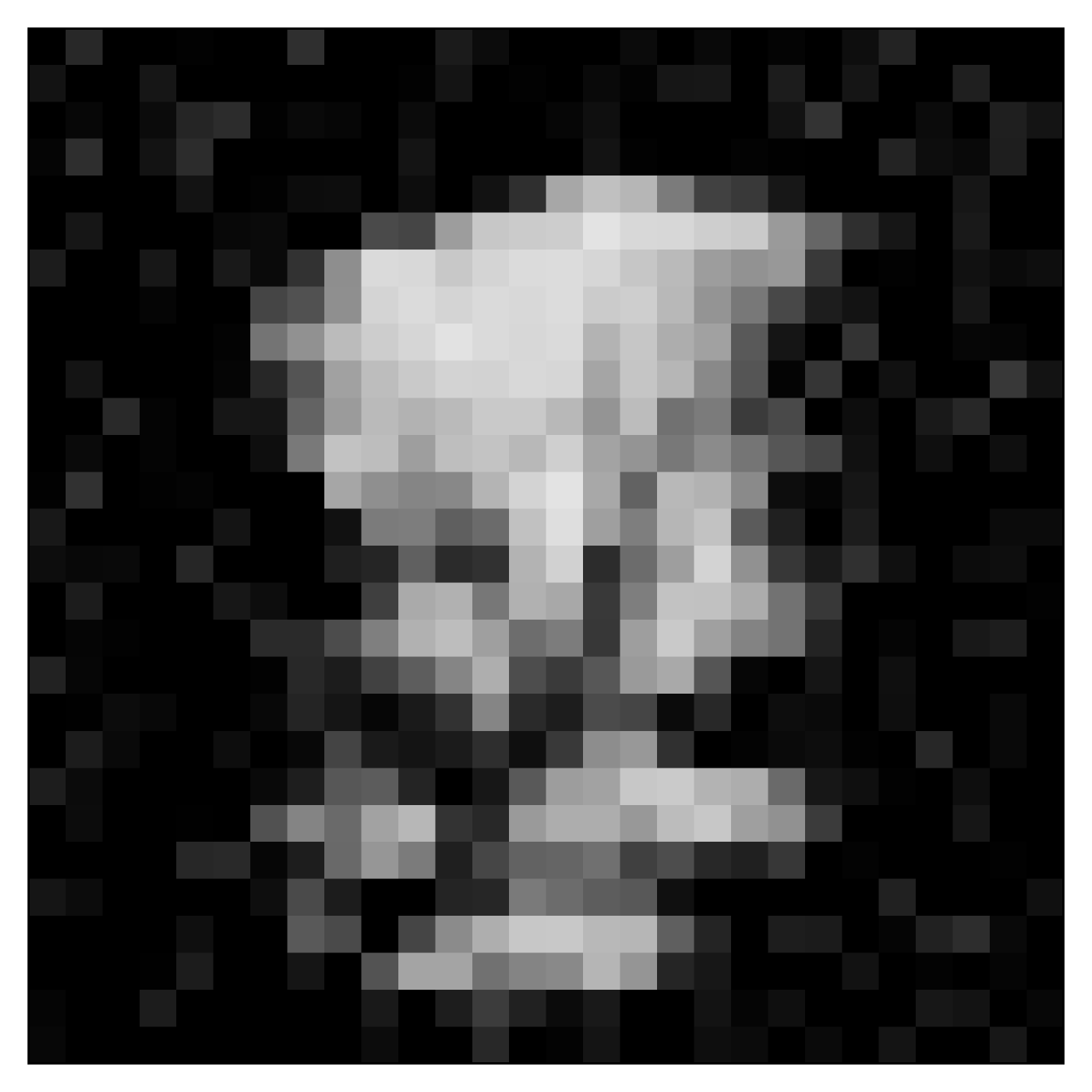}}
\\
\subfigure{\includegraphics[scale = 0.18]{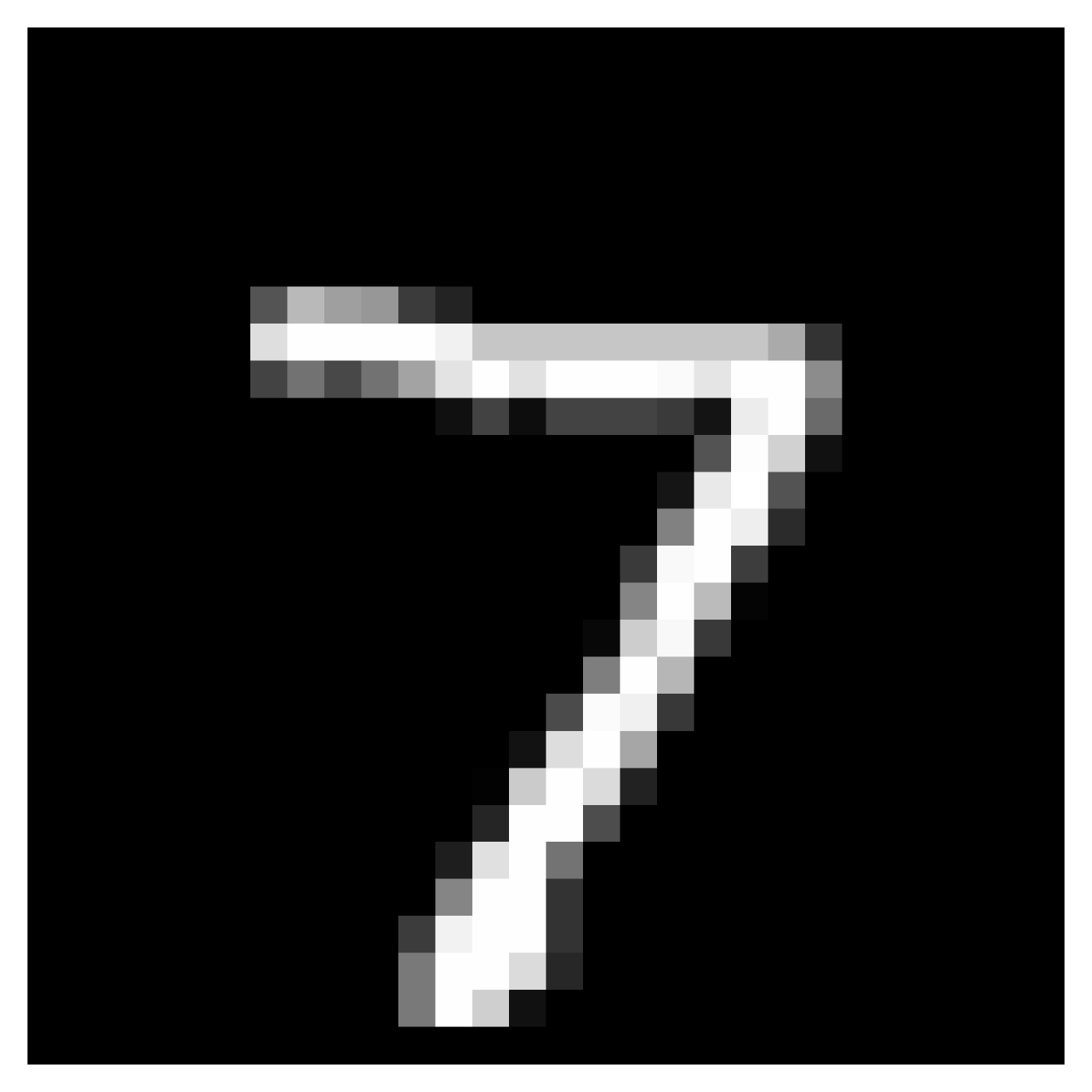}}
\subfigure{\includegraphics[scale = 0.18]{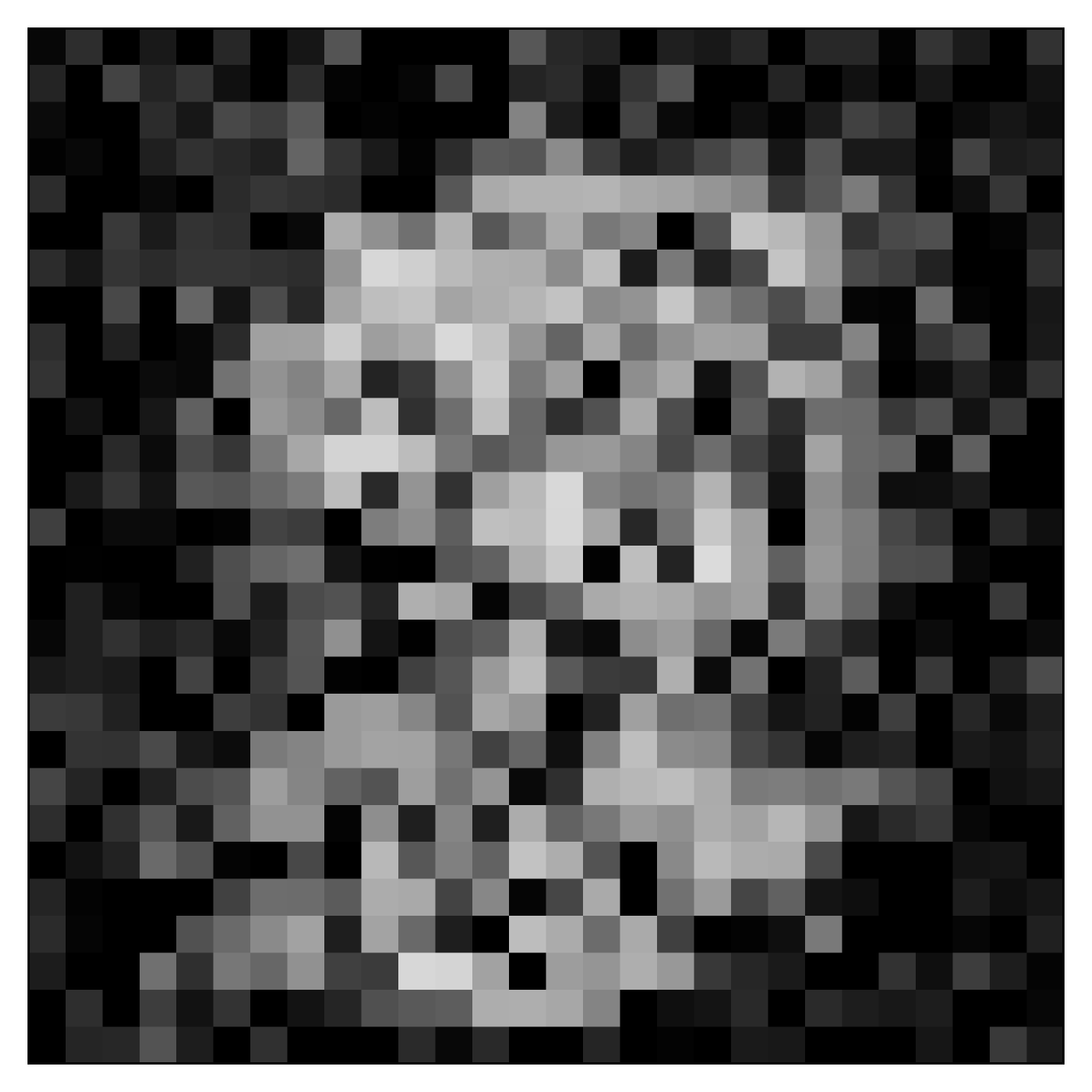}}
\subfigure{\includegraphics[scale = 0.18]{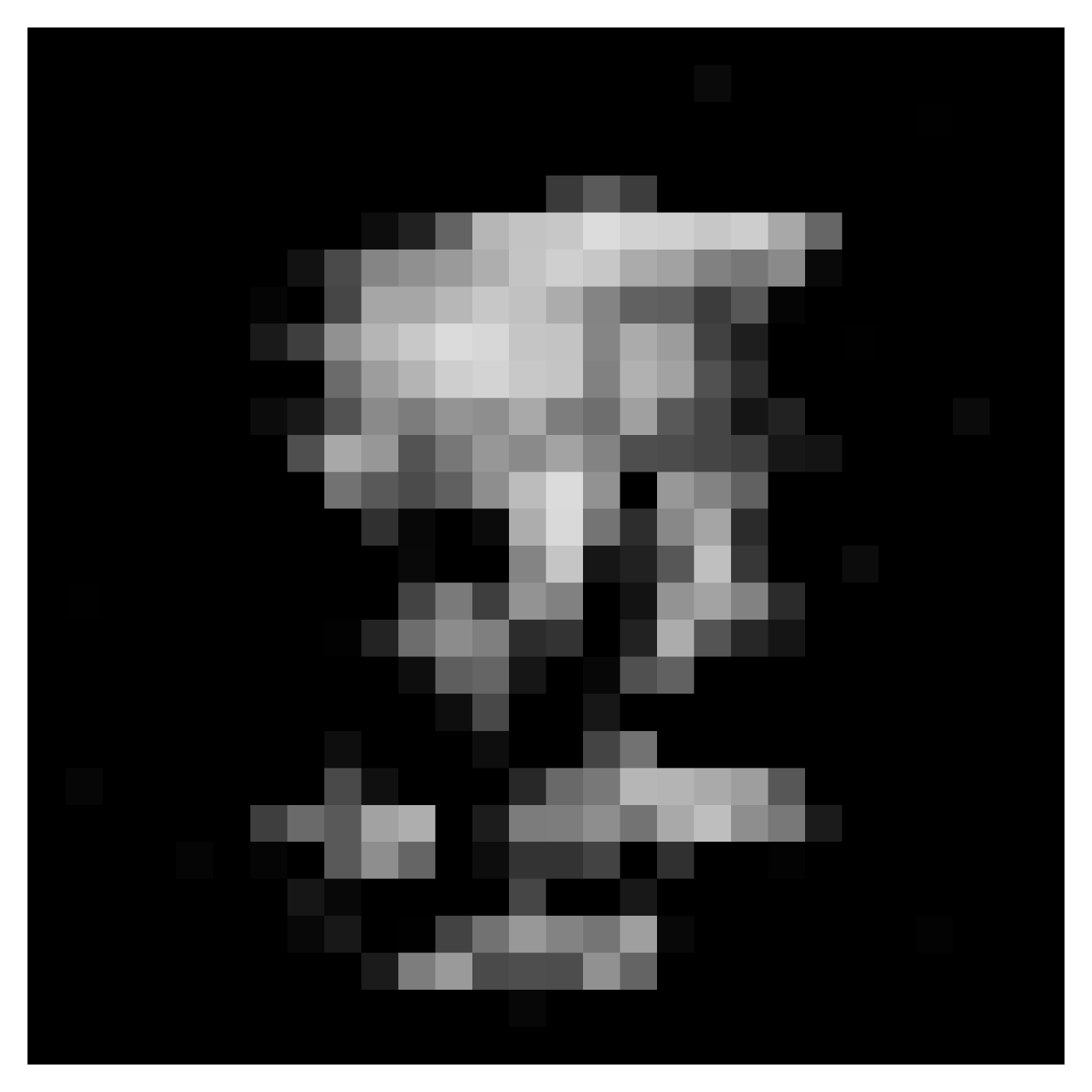}}
\caption{An example of bounding maps of images of digit `1' and `7'.
(Left) The original images, (Middle) Bounding map of a normal model. (Right) Bounding map of a robust model.} \label{fig:bounding_maps}
\end{figure}

To study the property of bounding maps, we take a simple example of binary classification: to distinguish digit `1' from `7' in MNIST dataset.
We use model with three hidden layers of 100 neurons and train it both in normal and adversarial way.
We visualize the bounding maps of both models for two example images in Figure \ref{fig:bounding_maps}.

For the normal model, the bounding maps are noisy and can hardly reveal the patterns of the input data.
However, for the robust model, the bounding maps can capture some intrinsic characteristics of the input data.
In the case of digit `1' and `7', people typically distinguish them by the horizontal stroke which digit `7' has and `1' does not.
This corresponds to the relatively smaller bounds of features in the middle above of the images.
It indicates the decision boundaries are closer to the data points in the directions of these features and the model puts more weight on these features to make predictions.
On the contrary, both digit `1' and `7' have a vertical stroke, we can correspondingly see a dark clear vertical stroke in the bounding maps of the robust model.
Such phenomenon can be reproduced in many other examples, more are available in Appendix \ref{sec:app_exp_interpret}.

We need to mention similar property of robust models is found in \citet{tsipras2018there} but from a more microscopic perspective.
\citet{tsipras2018there} visualizes the gradients of the model's loss function w.r.t the input data and finds that the gradients for robust models are \textit{significantly more interpretable}, while the gradients for normal models are generally noise.
Our investigations are more on a macroscopic level, our non-uniform bounds explore the shape of model's decision boundary but we have the same claim: \textit{robust models are more interpretable}.

\subsubsection{Robustness and Decision Boundary} \label{sec:geo_corr}

In Figure \ref{fig:bounding_maps}, similar patterns of bounding maps are found for the same model but different input images.
Thus, we calculate the cosine similarity of $\eps$ for two images, since the direction of $\eps$ indicates the shape of non-uniform bounds.
For examples of normal and robust models, we report the average and minimum values in all image pairs of the test set in Table \ref{tbl:cos_similarity_sim}.
The full results are available in Table \ref{tbl:app_cos_similarity} in Appendix \ref{sec:app_exp_sim}.
It is clear that the values of $\eps$ for different images but the same model are highly correlated, which indicates the geometric similarity of non-uniform bounds.
\footnote{For two random vectors uniformly distributed in $[0, 1]^{784}$, the expectation of cosine similarity between them is around $0.75$. The expectation decreases for random vectors in higher dimensions.}
What's more, such correlation is even stronger in the cases of robust models.

\begin{table}
\centering
\begin{tabular}{|c|c|c|}
\hline
& Mean Cosine & Minimum Cosine \\
\hline
Normal Model & 0.9774 & 0.5038 \\
\hline
Robust Model & 0.9964 & 0.9104 \\
\hline
\end{tabular}
\caption{The mean and minimum of cosine similarity between all pairs of $\eps$ in the test set.
We use MNIST classification model with 300 neurons in each hidden layer.} \label{tbl:cos_similarity_sim}
\end{table}

The high correlation means some features are consistently more robust than the other features in different input data points.
Since most $\eps$ are almost collinear, the direction of them can be regarded as a quantitative and data-agnostic metric measuring the robustness of input features.
It is also beneficial to use this direction as a prior when we estimate the non-uniform bound for a new data point.

Since the shape of the non-uniform bound reveals the decision boundary, high correlation of $\eps$ also indicates the uniformity of the direction of the decision boundary.
Formally, in a $n_1$ dimensional input space, there exists a subspace $\mathcal{X}$ of dimensionality $n'_1 \ll n_1$ that contains most directions of decision boundary around the data manifold.
This is consistent with what \citet{moosavi2017universal} points out.

An extreme example is the classifier whose decision boundary is linear, the non-uniform bound of the largest volume for any input data has exactly the same shape.
That is to say, the values of $\eps$ for any input data point are exactly collinear and $n'_1  = 1$ in this case.
Our experimental results show the stronger correlation of $\eps$ in the cases of robust models.
This implies the most directions of a robust model's decision boundary can be obtained in a subspace of even lower dimensionality than a normal model.
The decision boundary of a robust model should be simpler in some sense.

\section{Discussion} \label{sec:discussion}

In this section, two straightforward extensions are shown to make our algorithms adapt to other settings.
We also discuss the potential future works to polish the algorithms.

\subsection{Extensions}

\subsubsection{Other Network Architectures}

Some previous works have some assumptions on the architecture of the neural network.
For example, \citet{kolter2017provable} and \citet{weng2018towards} assume ReLU network; \citet{raghunathan2018certified} only works for network with only one hidden layer.
\citet{wong2018scaling} and \citet{zhang2018efficient} generalize the method of \citet{kolter2017provable} and \citet{weng2018towards} respectively to general feedforward networks.

Although our previous analysis in Section \ref{sec:alg} is based on the fully-connected network, our framework is modularized and can generalize naturally to general feedforward networks in the same way as \citet{wong2018scaling}.
Our method is from the primal perspective while \citet{wong2018scaling} focuses on the dual problem.

For example, the convolutional layers can be reparameterized as feedforward layers of sparse weight matrices and shared variables.
The max-pooling layers can be considered as non-linear functions and be linearized by methods in Section \ref{subsection:nonlinear_approx}.
Our framework can also be applied to network with shortcut connections, including popular Residual Network (ResNet) \cite{he2016deep} and Densely Connected Network (DenseNet) \cite{huang2017densely}.
More details about this are available in Appendix \ref{sec:app_architecture}.



\subsubsection{Other Norms} 

Much attention of previous works is focused on bounds based on $l_\infty$ norm \cite{zhang2018efficient, singh2018fast, kolter2017provable}, although some of them such as \citet{kolter2017provable} can be easily extended to attacks based on other norms.
We claim that our methods can be extended to other norms in the same way.
However, we also point out both previous works and this work implicitly favor $l_\infty$ norm when estimating the bounds of output logits.
This is because the neurons in intermediate layers are bounded in an \textit{elementwise} manner, the bound for a specific neuron does not depend on the bound of any other neuron.
This bound would be loose if the output $\z^{(i)}$, as a vector, can be better bounded by a norm other than $l_\infty$ norm.
It would be interesting to consider the bound of neurons in one layer jointly and derive a tighter bound for non-$l_\infty$ norms.

\subsection{Potential Future Works}

Solving problem (\ref{eq:problem}) exactly is difficult, because the exact range of output logits $\z^{(N)}$ when $\hat{\z}^{(1)} \in \Set_{\eps}(\x)$ is generally intractable.
Therefore, we introduce a tractable bound of $\z^{(N)}$ and build our algorithms based on that.
From the geometric perspective, the bound of $\z^{(N)}$ given by \textbf{Algorithm} \ref{alg:bound_est} provides a tractable envelope of the intractable decision boundary of the neural network.
The algorithm to bound $\z^{(N)}$ is important, because an algorithm of a tighter bound reveals the decision boundary better and then leads to certified regions of larger volumes.

One possible direction to explore is to design faster and better algorithms than \textbf{Algorithm} \ref{alg:bound_est}, which still has looser bounds for larger $\eps$ or deeper networks.
In addition, as \textbf{Algorithm} \ref{alg:bound_est} is called in every iteration when we optimize $\eps$ in \textbf{Algorithm} \ref{alg:al}, the complexity of the algorithm is also an issue.
All algorithms discussed in this paper calculate the `strict bounds' i.e. $\low^{(N)} \leq \z^{(N)} \leq \up^{(N)}$, it would be beneficial to design an algorithm for `soft bounds' i.e. $\tilde{\low}^{(N)} \lessapprox \z^{(N)} \lessapprox \tilde{\up}^{(N)}$ but with much faster speed.
We run `soft bounds' algorithm first to accelerate the optimization and run `strict bounds' algorithm at last to guarantee the hard constraints are satisfied.

Another direction towards certified regions of larger volumes and studying the decision boundary is to consider \textit{oblique bounds} instead of \textit{standard bounds}.
The only difference is to introduce an additional orthogonal matrix $\mathbf{A}$ to parameterize the adversarial budget $\Set^{(p)}_{\mathbf{A}, \eps}(\x) := \{\hat{\z}^{(1)} = \x + \mathbf{A}(\eps \odot \vi)| \|\vi\|_p = 1\}$.
More importantly, the elements in matrix $\mathbf{A}$ represents the correlation between different input features, which will give us more information about the shape of the decision boundary.

\section{Conclusion} \label{sec:conclusion}

In this paper, we study the certified non-uniform bounds around input data points.
We propose a general framework to estimate the output logits of different neural networks.
The goal of finding the bounds of the largest volumes is then formulated as a constrained optimization problem and we solve it by the augmented Langragian method.
Our experiments on synthetic data and real data show the non-uniform bounds have the larger volumes than uniform bounds.
In addition, we use our algorithm as a tool to explore the decision boundaries of different models.
Our results demonstrate at least three advantages of robust models:
1) the model relies on fewer features and has much larger certified non-uniform bounds;
2) the non-uniform bounds are significantly more interpretable;
3) the stronger geometric similarity of the non-uniform bounds gives a quantitative, data-agnostic metric of input features' robustness and implies a simpler decision boundary.

\section*{Acknowledgement}

We thank Po-An Wang for beneficial discussions.
We also thank Qi Dou and Ya-Ping Hsieh for their feedback on the initial manuscripts.
This work is supported by Microsoft Research and Chen Liu is funded by Microsoft Research PhD Scholarship Program.



\bibliography{references}
\bibliographystyle{icml2019}

\newpage

\onecolumn

\appendix

\section{Missing Algorithms}\label{sec:app_alg}

\subsection{Bound Estimation Algorithms}

Here we list the detailed pseudo code of \textit{quadratic} and \textit{simple} algorithms mentioned in Section \ref{subsection:bound_est}.
The complexity in matrix multiplications for \textbf{Algorithm} \ref{alg:complex_bound_est} and \ref{alg:naive_bound_est} and are $O(N^2)$ and $O(N)$ respectively.

\begin{minipage}{.49\textwidth}
    \begin{algorithm}[H]
    \small
     \begin{algorithmic}[1]
    \STATE Input: Parameters $\{\W^{(i)}, \bias^{(i)}\}_{i = 1}^{N - 1}$, perturbation set $\Set_{\eps}(\x)$.
    \STATE $\low^{(2)} = \W^{(1)}\x - \W^{(1)}_{+}\eps + \W^{(1)}_{-}\eps + \bias^{(1)}$
    \STATE $\up^{(2)} = \W^{(1)}\x - \W^{(1)}_{-}\eps + \W^{(1)}_{+}\eps + \bias^{(1)}$
    \STATE $\M^{(1)} = \W^{(1)}$
    \STATE $\phi^{(2)} = \W^{(1)}\x + \bias^{(1)}$
    \FOR {i = $2, ..., N -1$}
        \STATE Calculate $\D^{(i)}$, $\m_1^{(i)}$, $\m_2^{(i)}$ based on $\low^{(i)}$ and $\up^{(i)}$
        \STATE $\M^{(j)} = \W^{(i)}\D^{(i)}\M^{(j)}$ for $j = 1, ..., i - 1$
        \STATE $\M^{(i)} = \W^{(i)}$
        \STATE $\phi^{(i + 1)} = \W^{(i)}\D^{(i)}\phi^{(i)} + \bias^{(i)}$
        \STATE $\low^{(i + 1)} = \phi^{(i + 1)} + \sum_{j = 1}^{i} \left( \M^{(j)}_{-}\m_2^{(j)} + \M^{(j)}_{+}\m_1^{(j)} \right)$
        \STATE $\up^{(i + 1)} = \phi^{(i + 1)} + \sum_{j = 1}^{i} \left( \M^{(j)}_{-}\m_1^{(j)} + \M^{(j)}_{+}\m_2^{(j)} \right)$
    \ENDFOR
    \STATE Output: Bounds $\{\low^{(i)}, \up^{(i)}\}_{i = 2}^{N}$
    \end{algorithmic}
   \caption{\textit{Quadratic} Bound Estimation} \label{alg:complex_bound_est}
    \end{algorithm}
\end{minipage}
\begin{minipage}{.49\textwidth}
    \begin{algorithm}[H]
    \small
    \begin{algorithmic}[1]
    \STATE Input: Parameters $\{\W^{(i)}, \bias^{(i)}\}_{i = 1}^{N - 1}$, perturbation set $\Set_{\eps}(\x)$.
    \STATE $\low^{(2)} = \W^{(1)}\x - \W^{(1)}_{+}\eps + \W^{(1)}_{-}\eps + \bias^{(1)}$
    \STATE $\up^{(2)} = \W^{(1)}\x - \W^{(1)}_{-}\eps + \W^{(1)}_{+}\eps + \bias^{(1)}$
    \FOR {i = $2, ..., N - 1$}
        \STATE $\low^{(i)} = \sigma(\low^{(i)})$
        \STATE $\up^{(i)} = \sigma(\up^{(i)})$
        \STATE $\low^{(i + 1)} = \W^{(i + 1)}_{+}\low^{(i)} + \W^{(i + 1)}_{-}\up^{(i)} + \bias^{(i)}$
        \STATE $\up^{(i + 1)} = \W^{(i + 1)}_{-}\low^{(i)} + \W^{(i + 1)}_{+}\up^{(i)} + \bias^{(i)}$
    \ENDFOR
    \STATE Output: Bound $\{\low^{(i)}, \up^{(i)}\}_{i = 2}^{N}$
    \end{algorithmic}
    \caption{\textit{Simple} Bound Estimation} \label{alg:naive_bound_est}
    \end{algorithm}
    \vspace{1.44cm}
\end{minipage}

\subsection{Gradient Calculation}

Below is the missing algorithm to calculate the gradient of bounds $\{\low^{(i)}, \up^{(i)}\}_{i = 2}^{N}$ w.r.t. $\eps$.
We put $g$ in front of a variable to represent its gradient w.r.t. $\eps$.
Terms like $g[\M]_{+}$ or $g[\M]_{-}$ are a bit abused here.
It means we put the elements of $g\M$ where the corresponding elements in $\M$ are positive or negative and set the other elements to be $0$.
In addition, $f_\mathbf{1}$ is the indicator function which returns $1$ if input is true and $0$ otherwise.
When the input is a tensor, the function is applied elementwisely and return a tensor of the same shape.

\begin{algorithm}
\small
\begin{algorithmic}[1]
\STATE Input: Parameters $\{\W^{(i)}, \bias^{(i)}\}_{i = 1}^{N - 1}$, perturbation set $\Set_{\eps}(\x)$, bounds $\{\low^{(i)}, \up^{(i)}\}_{i = 2}^N$, values $\{\D^{(i)}, \m_1^{(i)}, \m_2^{(i)}\}_{i = 1}^N$
\STATE $g\low^{(2)} = -|\W^{(1)}|$
\STATE $g\up^{(2)} = |\W^{(1)}|$
\STATE $g\M^{(1)} = \mathbf{0}$
\STATE $g\phi^{(2)} = 0$
\FOR {i = $2, ..., N - 1$}
    \STATE Run algorithm \ref{alg:bound_est} to obtain $\D^{(i)}$, $\m_1^{(i)}$ and $\m_2^{(i)}$.
    \STATE $\low^{(i + 1)}_{naive} = \W^{(i)}_{+}\left[\sigma'(\low^{(i)}) \odot g\low^{(i)}\right] + \W^{(i)}_{-}\left[\sigma'(\up^{(i)}) \odot g\up^{(i)}\right]$
    \STATE $\up^{(i + 1)}_{naive} = \W^{(i)}_{-}\left[\sigma'(\low^{(i)}) \odot g\low^{(i)}\right] + \W^{(i)}_{+}\left[\sigma'(\up^{(i)}) \odot g\up^{(i)}\right]$
    \STATE $g\D^{(i)} = g\low^{(i)}\frac{\partial \D^{(i)}}{\partial \low^{(i)}} + g\up^{(i)}\frac{\partial \D^{(i)}}{\partial \up^{(i)}}$
    \STATE $g\m_1^{(i)} = g\low^{(i)}\frac{\partial \m_1^{(i)}}{\partial \low^{(i)}} + g\up^{(i)}\frac{\partial \m_1^{(i)}}{\partial \up^{(i)}}$
    \STATE $g\m_2^{(i)} = g\low^{(i)}\frac{\partial \m_2^{(i)}}{\partial \low^{(i)}} + g\up^{(i)}\frac{\partial \m_2^{(i)}}{\partial \up^{(i)}}$
    \STATE $g\M^{(j)} = \W^{(i)}g\D^{(i)}\M^{(j)} + \W^{(i)}\D^{(i)}g\M^{(j)}$ for $j = 1, ..., i - 1$
    \STATE $g\M^{(i)} = \mathbf{0}$
    \STATE $g\phi^{(i + 1)} = \W^{(i)}g\D^{(i)}\phi^{(i)} + \W^{(i)}\D^{(i)}g\phi^{(i)}$
    \STATE $g\low^{(i + 1)}_{comp} = \sum_{j = 1}^i \left( g[\M^{(j)}]_{-}\m_2^{(j)} + [\M^{(j)}]_{-}g\m_2^{(j)} + g[\M^{(j)}]_{+}\m_1^{(j)} + [\M^{(j)}]_{+}g\m_1^{(j)} \right) + g\phi^{(i + 1)}$
    \STATE $g\up^{(i + 1)}_{comp} = \sum_{j = 1}^i \left( g[\M^{(j)}]_{-}\m_1^{(j)} + [\M^{(j)}]_{-}g\m_1^{(j)} + g[\M^{(j)}]_{+}\m_2^{(j)} + [\M^{(j)}]_{+}g\m_2^{(j)} \right) + g\phi^{(i + 1)}$
    \STATE $g\low^{(i + 1)} = f_\mathbf{1}(\low^{(i + 1)}_{naive} > \low^{(i + 1)}_{comp}) \odot g\low^{(i + 1)}_{naive} + f_\mathbf{1}(\low^{(i + 1)}_{naive} \leq \low^{(i + 1)}_{comp}) \odot g\low^{(i + 1)}_{comp}$
    \STATE $g\up^{(i + 1)} = f_\mathbf{1}(\up^{(i + 1)}_{naive} \leq \up^{(i + 1)}_{comp}) \odot g\up^{(i + 1)}_{naive} + f_\mathbf{1}(\up^{(i + 1)}_{naive} > \up^{(i + 1)}_{comp}) \odot g\up^{(i + 1)}_{comp}$
\ENDFOR
\STATE Output: Gradients $\{g\low^{(i)}, g\up^{(i)}\}_{i = 2}^{N}$
\end{algorithmic}
\caption{Gradient Calculation}\label{alg:grad_calculation}
\end{algorithm}

In \textbf{Algorithm} \ref{alg:grad_calculation}, $g\D^{(i)}$, $g\m_1^{(i)}$ and $g\m_2^{(i)}$ can be obtained immediately after the calculation of $\D^{(i)}$, $\m_1^{(i)}$ and $\m_2^{(i)}$.
We can run the for-loop in \textbf{Algorithm} \ref{alg:grad_calculation} immediately after the corresponding iteration in \textbf{Algorithm} \ref{alg:bound_est}.
That is to say, we do not need to wait for \textbf{Algorithm} \ref{alg:bound_est} to terminate before calculating the gradients like what back-propagation does.
This can improve the computational efficiency.

\section{Different Bound Estimation Algorithms} \label{sec:app_discussion}

\begin{figure}[H]
\centering
\subfigure[Example Network 1]{
    \includegraphics[scale = 0.5]{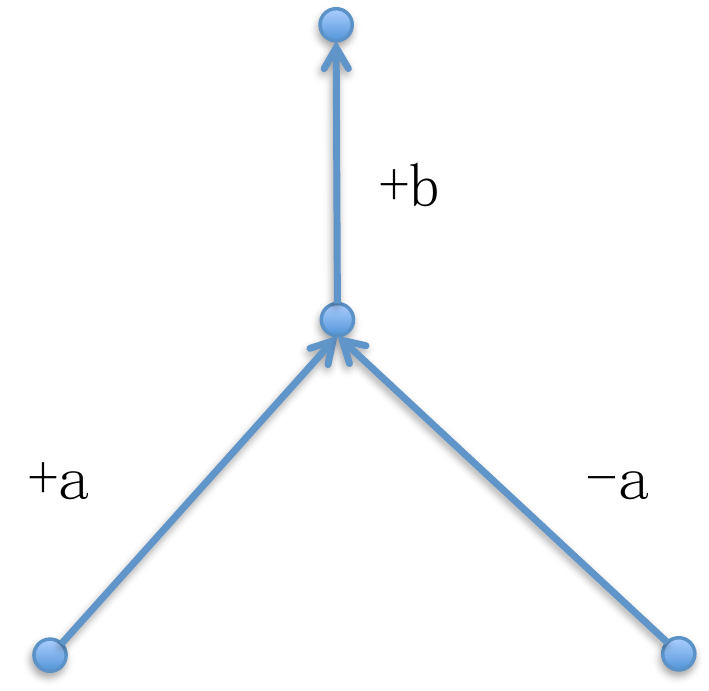} \label{fig:example_net1}
}
~~~~~~~~
\subfigure[Example Network 2]{
    \includegraphics[scale = 0.5]{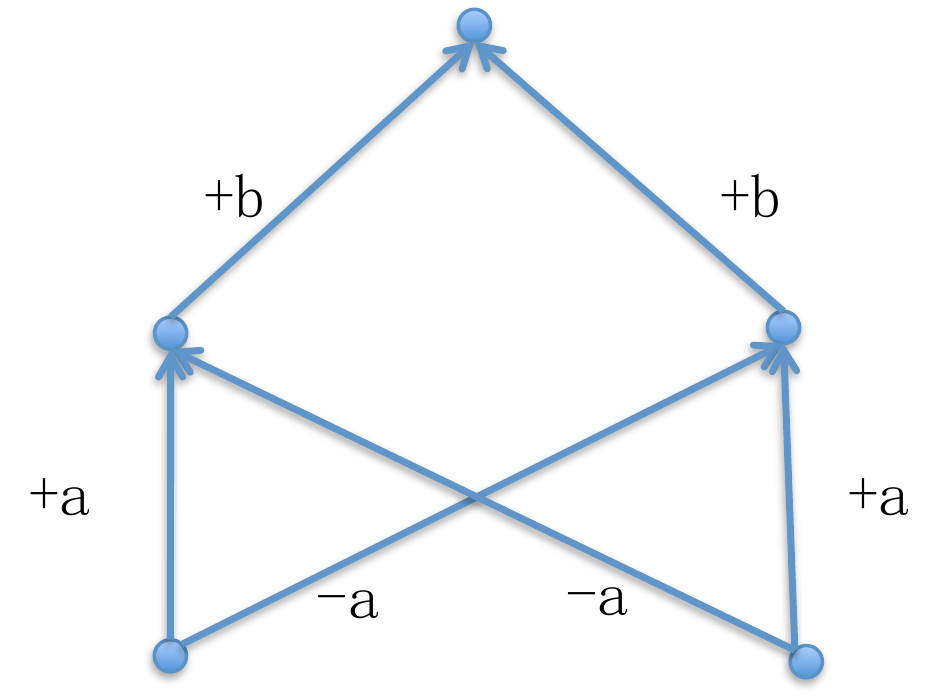} \label{fig:example_net2}
}
\caption{Two toy networks.} \label{fig:example_nets}
\end{figure}

It is a bit counterintuitive to find that \textbf{Algorithm} \ref{alg:complex_bound_est} and \ref{alg:naive_bound_est} are actually complementary.
There is \textit{no guarantee} which one is better in all cases and combining them together in \textbf{Algorithm} \ref{alg:bound_est} is a necessary.
We use the following two toy examples in Figure \ref{fig:example_nets} to demonstrate the pros and cons of both algorithms.

The activation function in the hidden layer is ReLU, the weights of each connection are shown in Figure \ref{fig:example_nets} and we assume $a, b > 0$.
Let the input point be $(x, x)$ and $x > 0$.
The adversarial budgets for both features are $\epsilon > 0$.
We consider the bounds of both algorithms for both networks as follows.

For network 1, the bounds of the pre-activation for the only hidden neuron are $[-2a\epsilon, +2a\epsilon]$ in both algorithms.
\textit{Simple algorithm} obtains $[0, +2a\epsilon]$ as the bound for post-activation and $[0, +2ab\epsilon]$ for the final output.
On the other hand, \textit{quadratic algorithm} obtains $\D^{(2)} = \left[+0.5\right]$ and $\left[0\right]^T \leq \m^{(2)} \leq \left[+a\epsilon\right]^T$.
Then the expression for the final output is $\left[+b\right]\left[+0.5\right]\begin{bmatrix}+ a & - a\end{bmatrix} \m^{(1)} + \left[+b\right] \m^{(2)} = \begin{bmatrix}+0.5ab & -0.5ab\end{bmatrix}\m^{(1)} + \left[+b\right] \m^{(2)}$.
As $\begin{bmatrix}-\epsilon & -\epsilon\end{bmatrix}^T \leq \m^{(1)} \leq \begin{bmatrix}+\epsilon & +\epsilon\end{bmatrix}^T$ given by perturbation budget, so the final output bounds of \textit{quadratic algorithm} are $[-ab\epsilon, +2ab\epsilon]$.
The true bound for the output in network 1 is $[0, +2ab\epsilon]$, so \textit{simple algorithm} wins in this case.

Similarly, the bounds of the pre-activation for both neurons in hidden layer in network 2 are $[-2a\epsilon, +2a\epsilon]$.
\textit{Simple algorithm} obtains $[0, +2a\epsilon]$ as the bound for post-activation and $[0, +4ab\epsilon]$ for the final output.
On the other hand, \textit{quadratic algorithm} obtains $\D^{(2)} = diag([+0.5, +0.5])$ and $[0, 0]^T \leq \m^{(2)} \leq [+ a\epsilon, + a\epsilon]^T$.
The expression for the final output is $\begin{bmatrix}+b & +b\end{bmatrix}\begin{bmatrix}+0.5 & 0\\0 & +0.5\end{bmatrix}\begin{bmatrix}+a & -a \\ -a & +a\end{bmatrix}\m^{(1)} + \begin{bmatrix}+b & +b\end{bmatrix}\m^{(2)} = \begin{bmatrix}+b & +b\end{bmatrix}\m^{(2)}$ and the final bound is $[0, +2ab\epsilon]$.
The true bound for the output in network 2 is also $[0, +2ab\epsilon]$, so \textit{quadratic algorithm} wins in this case.

Obviously, as the combination of both algorithms, \textbf{Algorithm} \ref{alg:bound_est} outputs the optimal in both cases.

To summarize, \textit{quadratic algorithm} can, to some extent, capture the composition of transformations in the neural network.
For example, the pre-activations of the hidden layer in network 2 are always additive inverse and this constrains the output range of the network.
\textit{Quadratic algorithm} can detect this constraint when calculating the output bound, as the first term of the final output's expression cancels out.
\textit{Simple algorithm} totally ignores that as it will discard all information of previous layers not directly connected to the current layer.
However, \textit{quadratic algorithm} use a linear approximation of the activation function while \textit{simple algorithm} use the exact one.
Unnecessary linearization made \textit{quadratic algorithm} obtain suboptimal bound in cases like network 1.

Empirically, \textit{simple algorithm} prefers larger perturbation, because the linearization of activation function invokes larger error here.
\textit{Quadratic algorithm} is suitable when the layer size is large, as the composition of large matrix transformation typically means more terms can be cancelled out.

\begin{figure}[H]
\centering
\subfigure[2-5-5-2 network, $\epsilon$ = 0.1]{
   \includegraphics[scale = 0.33]{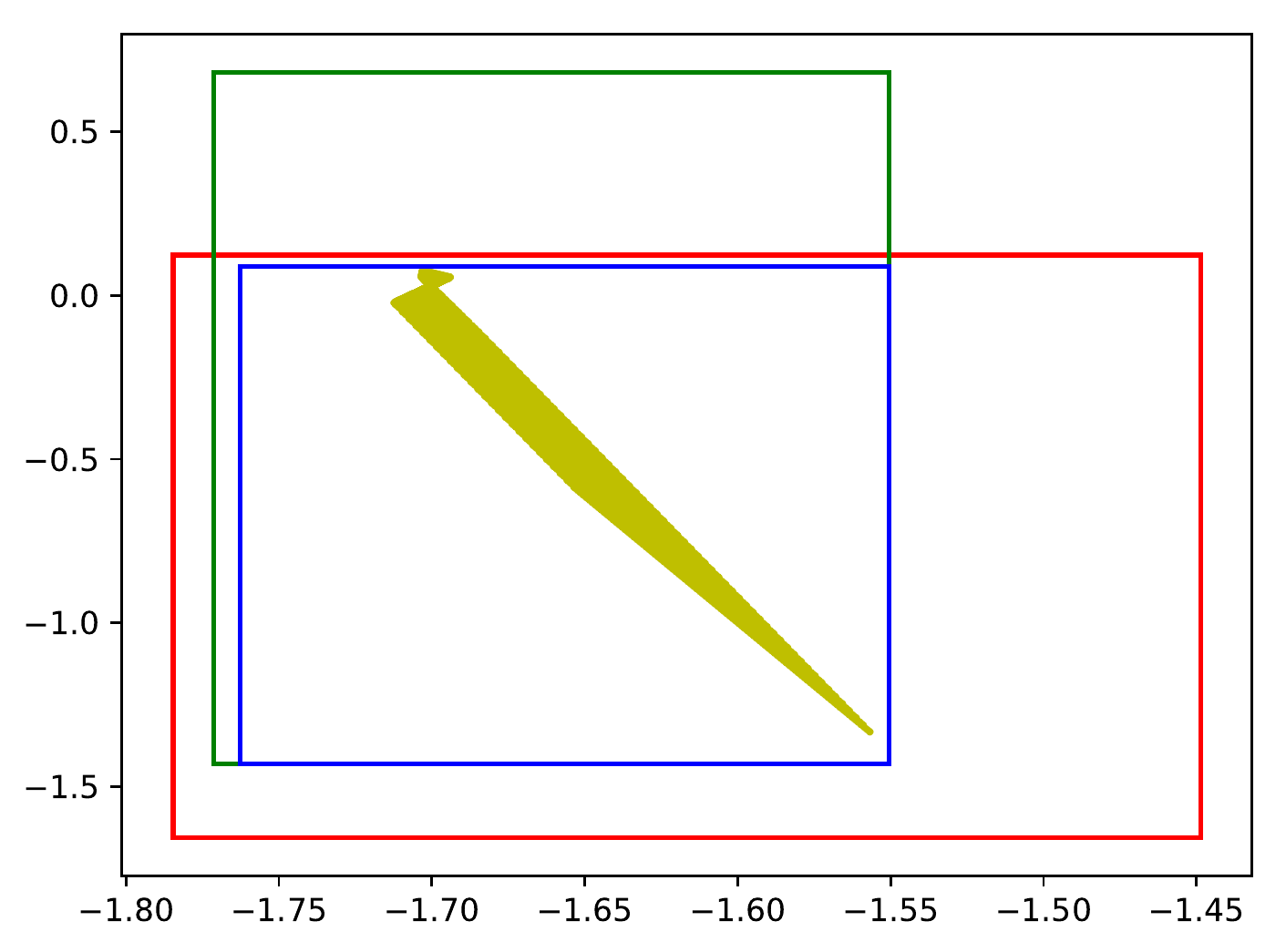}
}
~~~
\subfigure[2-10-10-2 network, $\epsilon$ = 0.1]{
    \includegraphics[scale = 0.33]{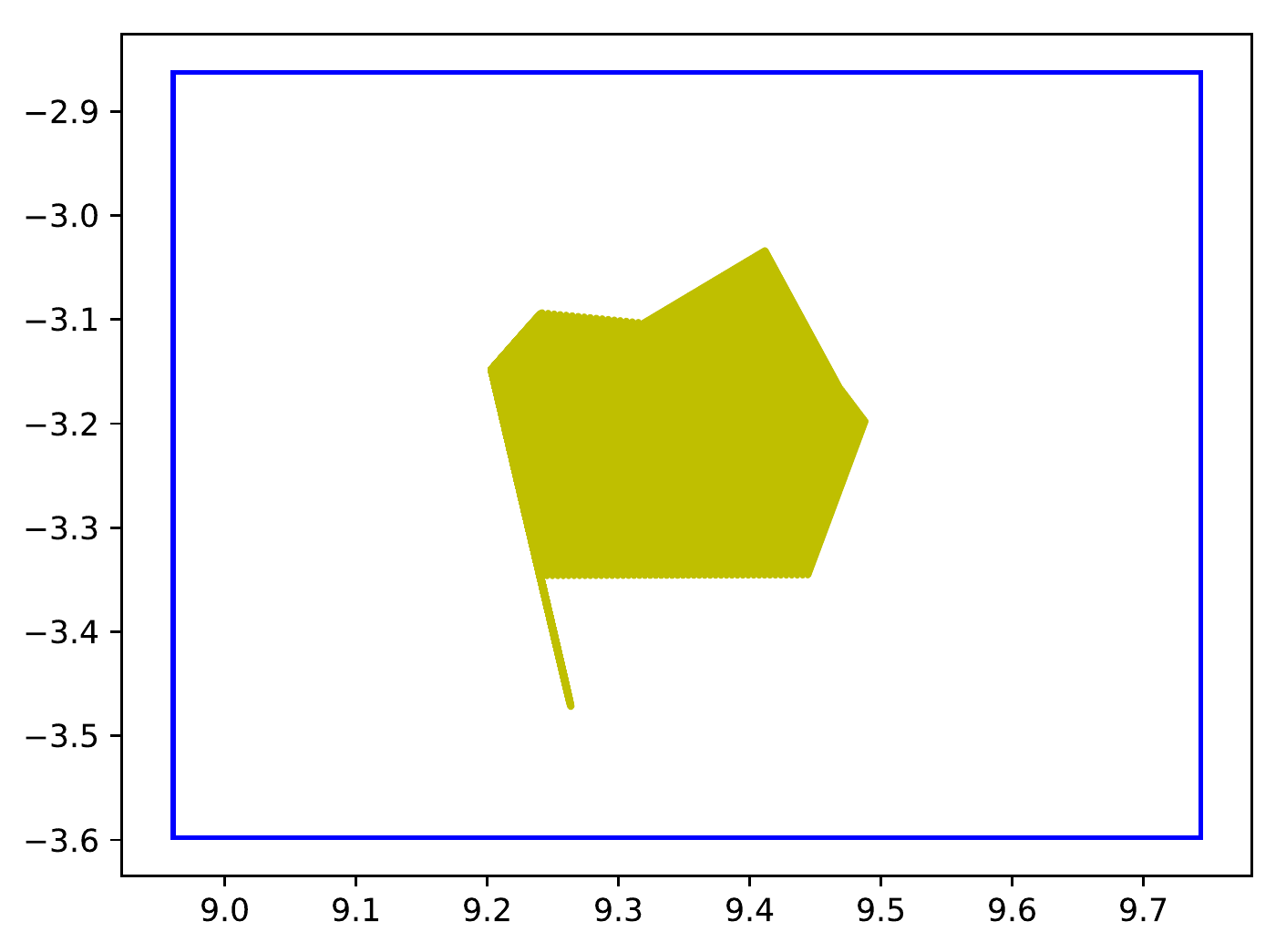}
}
~~~
\subfigure[2-10-10-2 network, $\epsilon$ = 0.2]{
    \includegraphics[scale = 0.33]{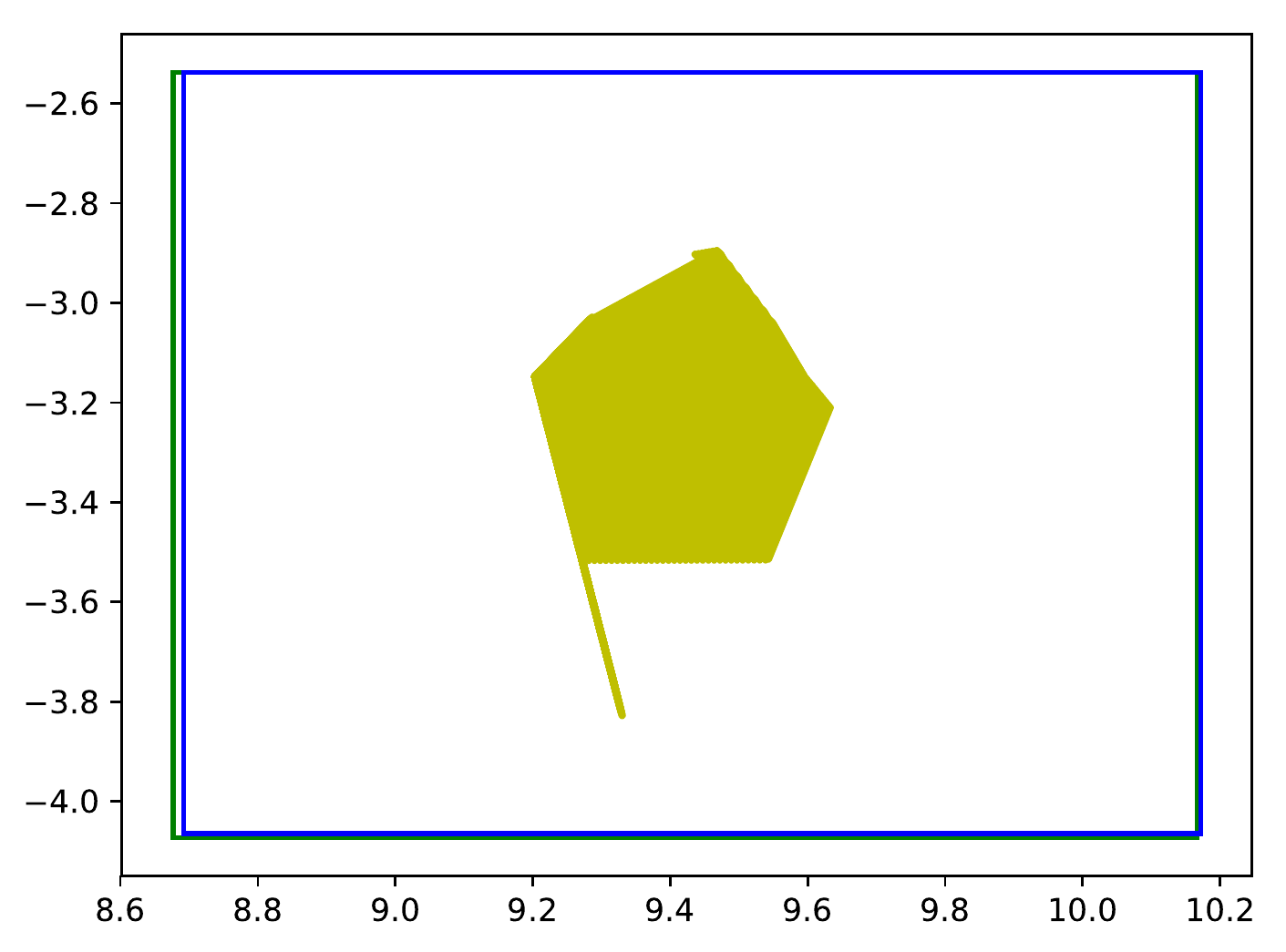}
}
\caption{Visualization of bounds from \textit{simple} (red), \textit{quadratic} (green) and combined (blue) algorithms with different hidden neurons and uniform bounded perturbation budgets $\tau$. (bounds of \textit{simple algorithm} for (b) and (c) are out of scope.) The set of all possible outputs are in yellow.} \label{fig:bounds_compare_example}
\end{figure}

Figure \ref{fig:bounds_compare_example} demonstrates the different bounds and possible outputs in different toy cases.
We generate neural networks with two hidden layers of random weights and compare the bounds obtained by different algorithms.
In these cases, we can see the bounds of \textit{simple algorithm} become loose quickly with the increase of layer size while the bounds of \textit{quadratic algorithm} might be the worse when perturbation budgets are large.
Of course, as \citet{kolter2017provable} points out, all bounds become looser in larger layer size and larger perturbation budgets.

\section{Extra Experiment Results}

\subsection{Volume of Bounds} \label{sec:app_exp_general}

\begin{figure}
\subfigure[`100-100-100' models on MNIST]{\includegraphics[scale = 0.55]{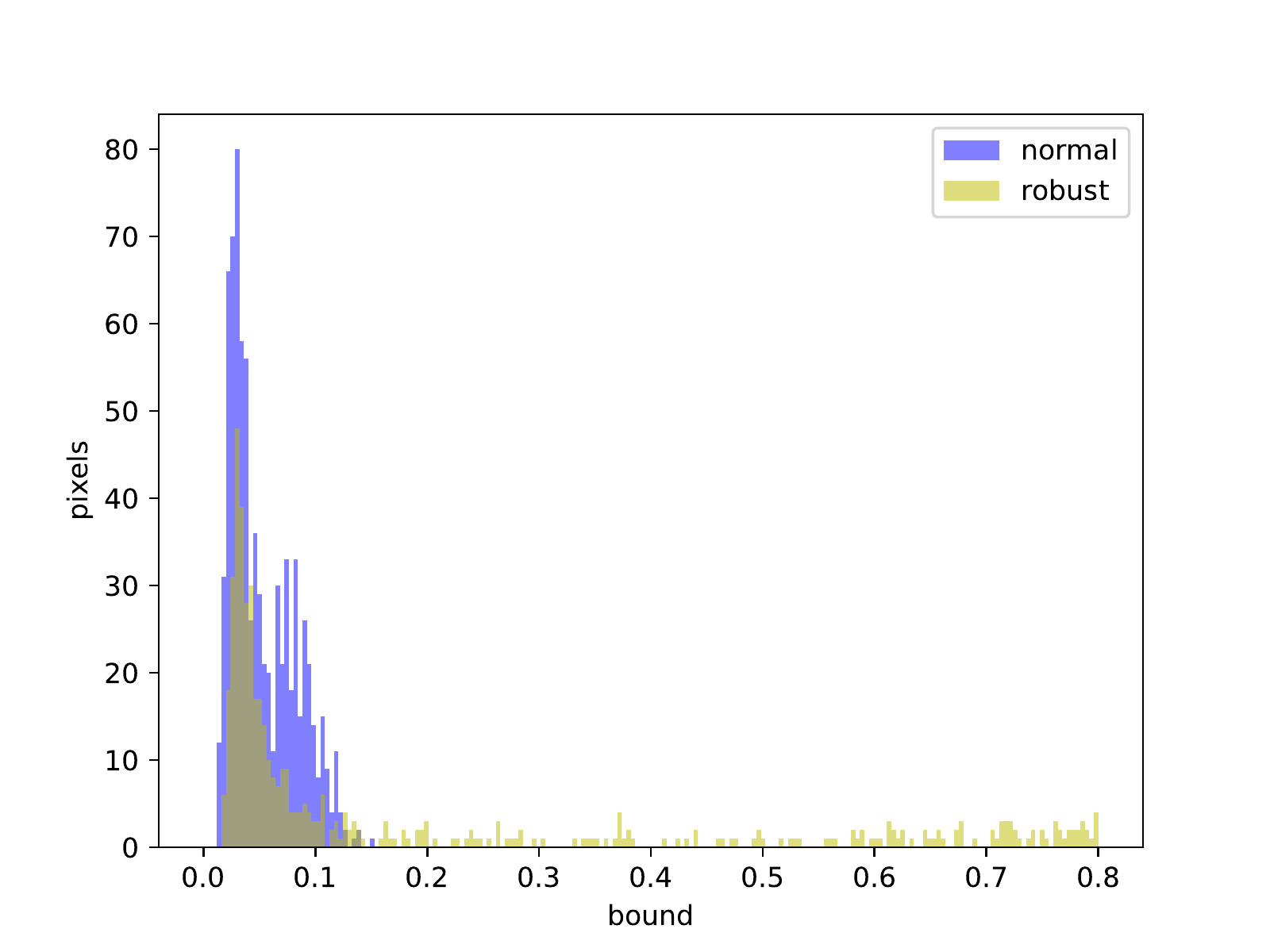}}
\subfigure[`500-500-500' models on MNIST]{\includegraphics[scale = 0.55]{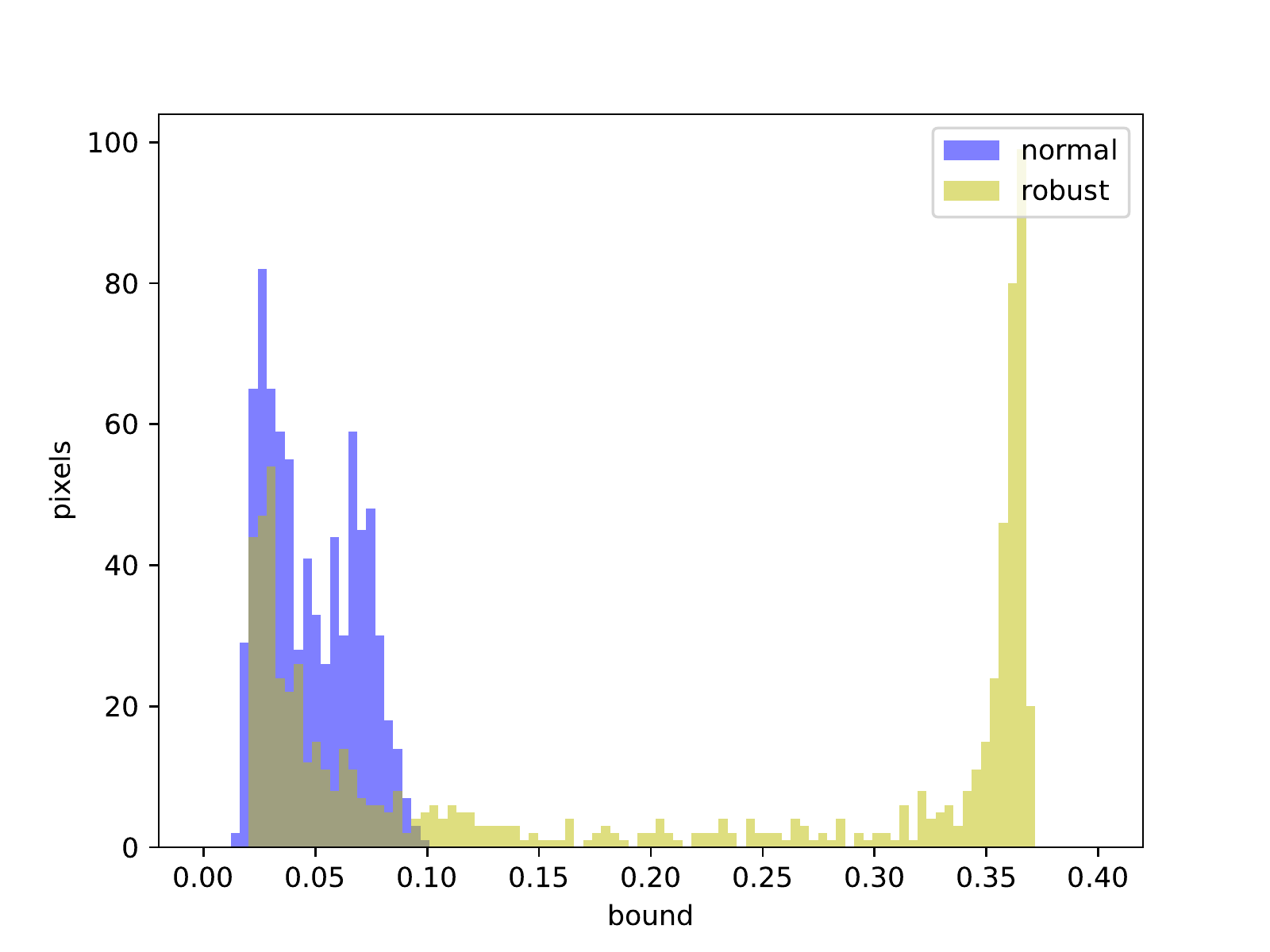}} \\
\subfigure[`1024-1024-1024' models on Fashion-MNIST]{\includegraphics[scale = 0.55]{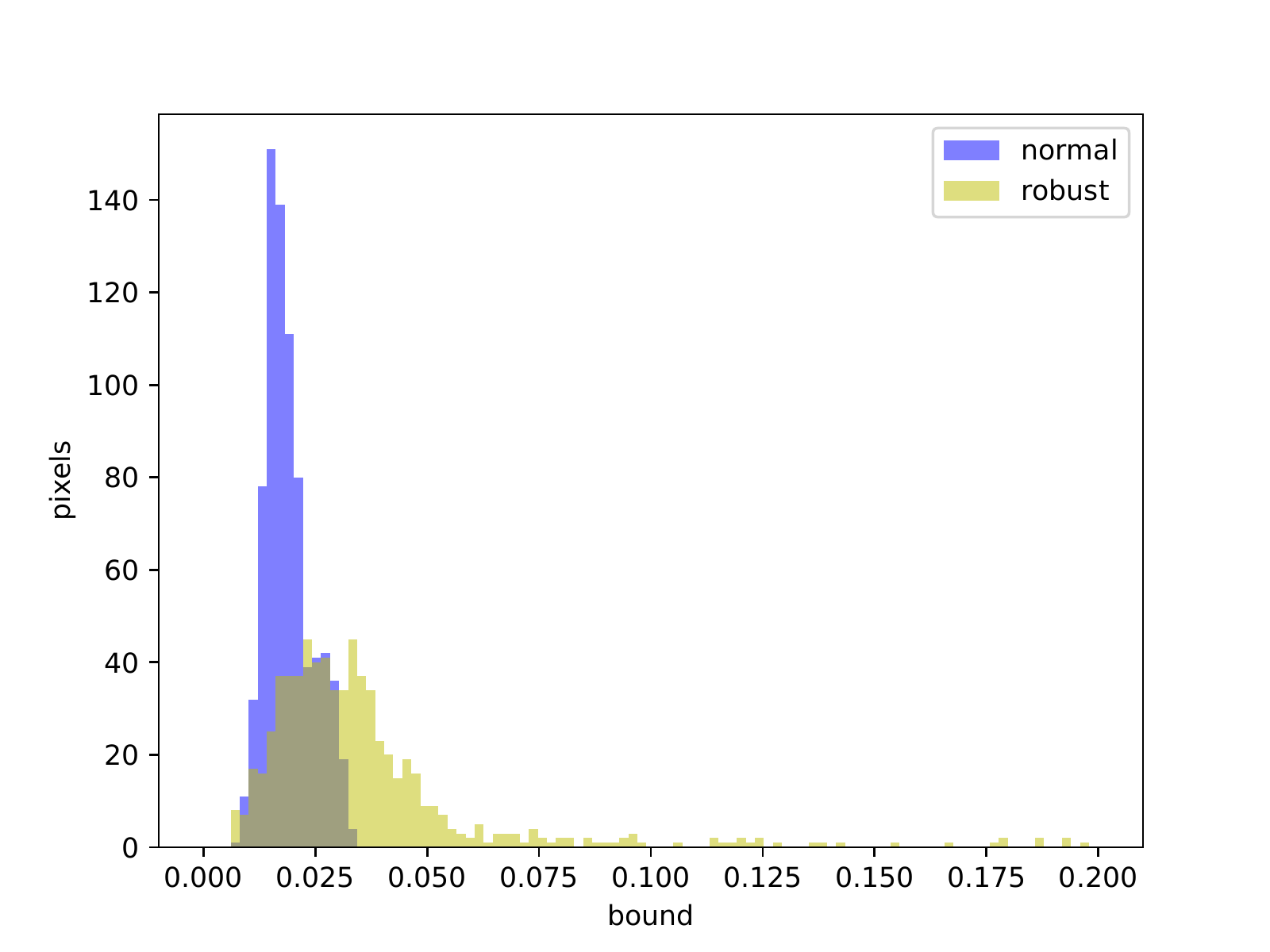}}
\subfigure[`1024-1024-1024' models on SVHN]{\includegraphics[scale = 0.55]{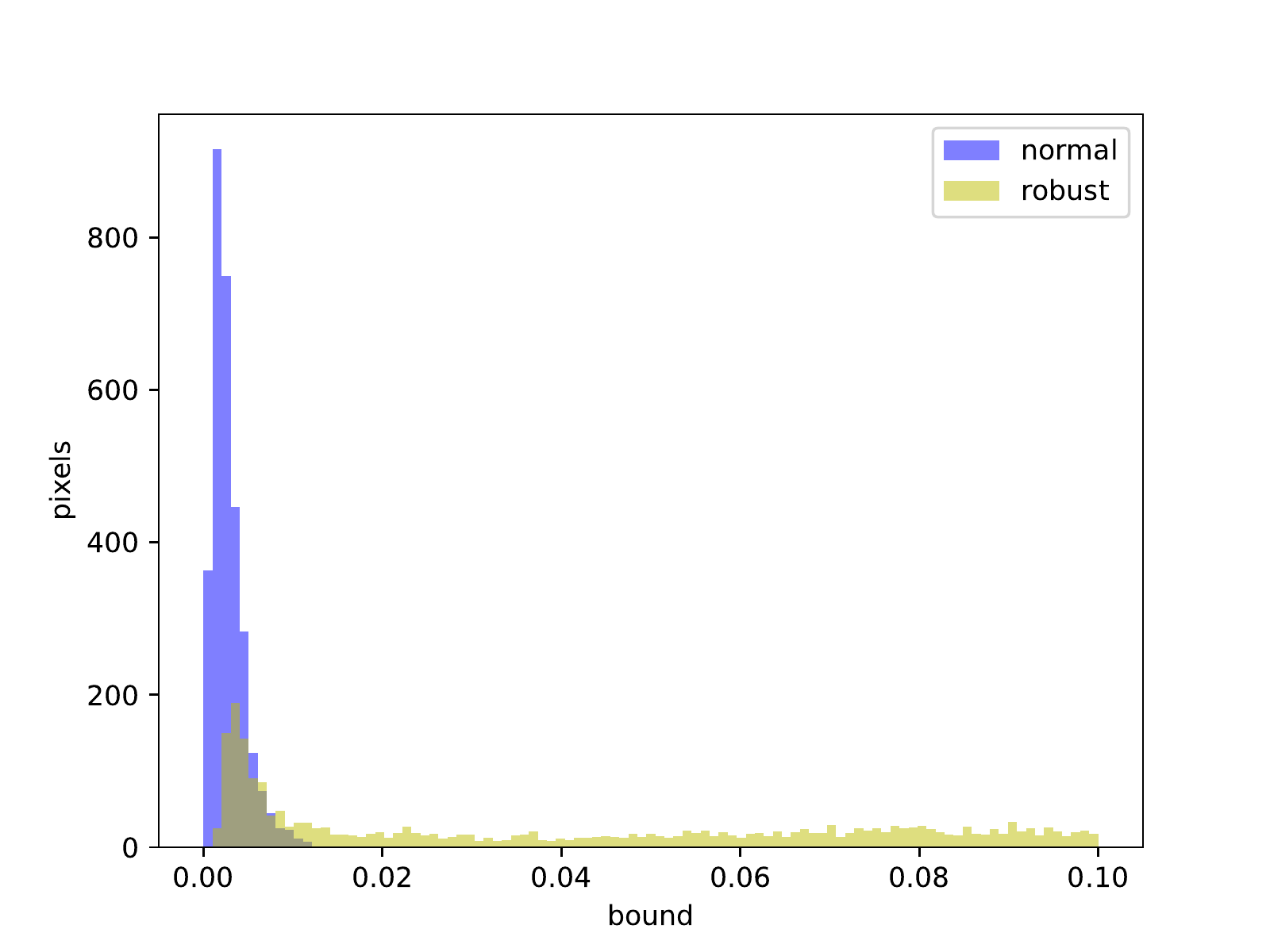}}
\caption{More examples of distributions of bounds for normal and robust models among all pixels.} \label{fig:app_hist_bounds}
\end{figure}

Like Figure \ref{fig:hist_bounds}, we put examples of other models and other datasets in Figure \ref{fig:app_hist_bounds}.
We show that the results are consistent: the distributions of non-uniform bounds among input features of the robust models have a `long tail', which indicates the output logits are affected little by some input features.



\subsection{Robustness and Model Interpretability} \label{sec:app_exp_interpret}

\begin{figure}[h!]
\subfigure{\includegraphics[scale = 0.18]{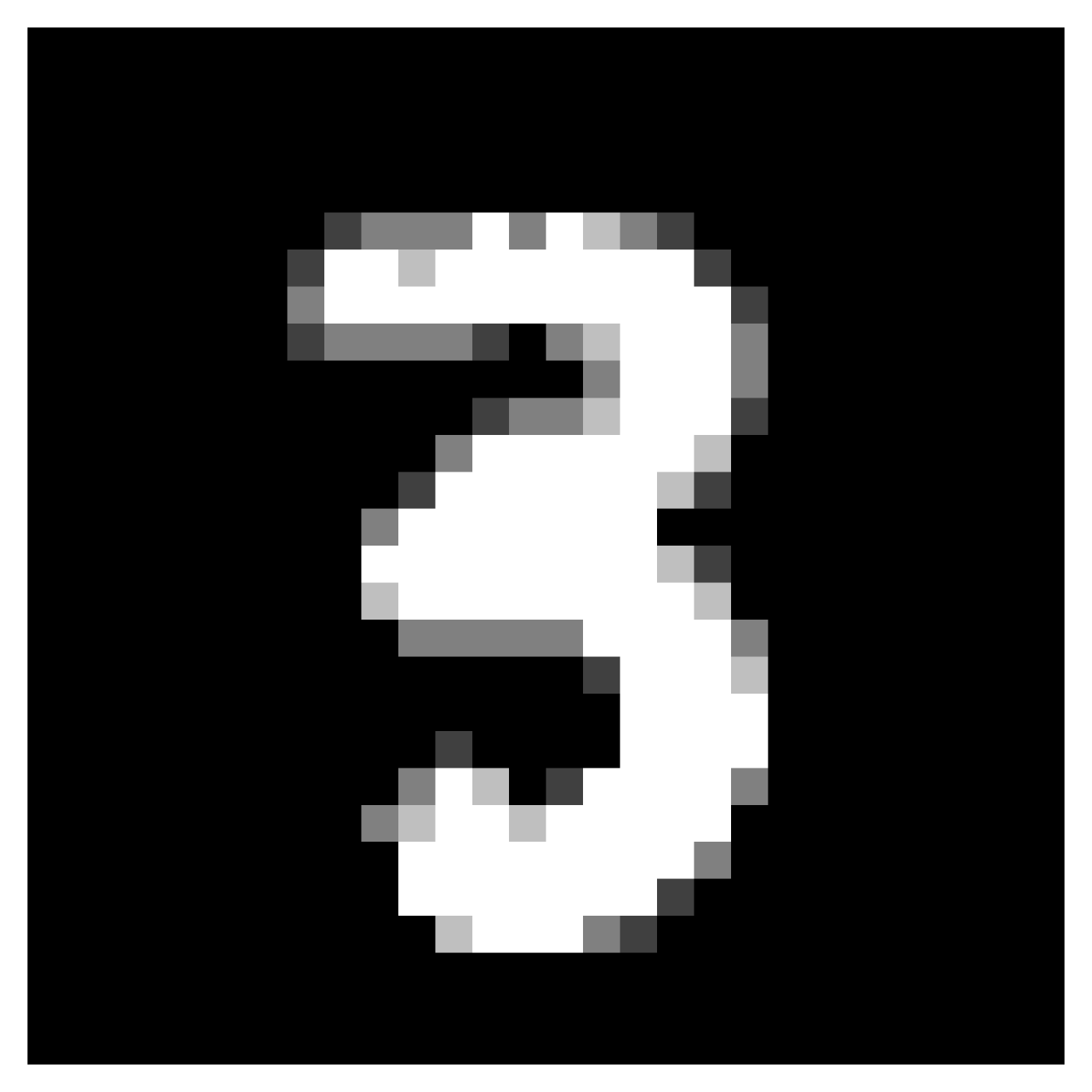}}
\subfigure{\includegraphics[scale = 0.18]{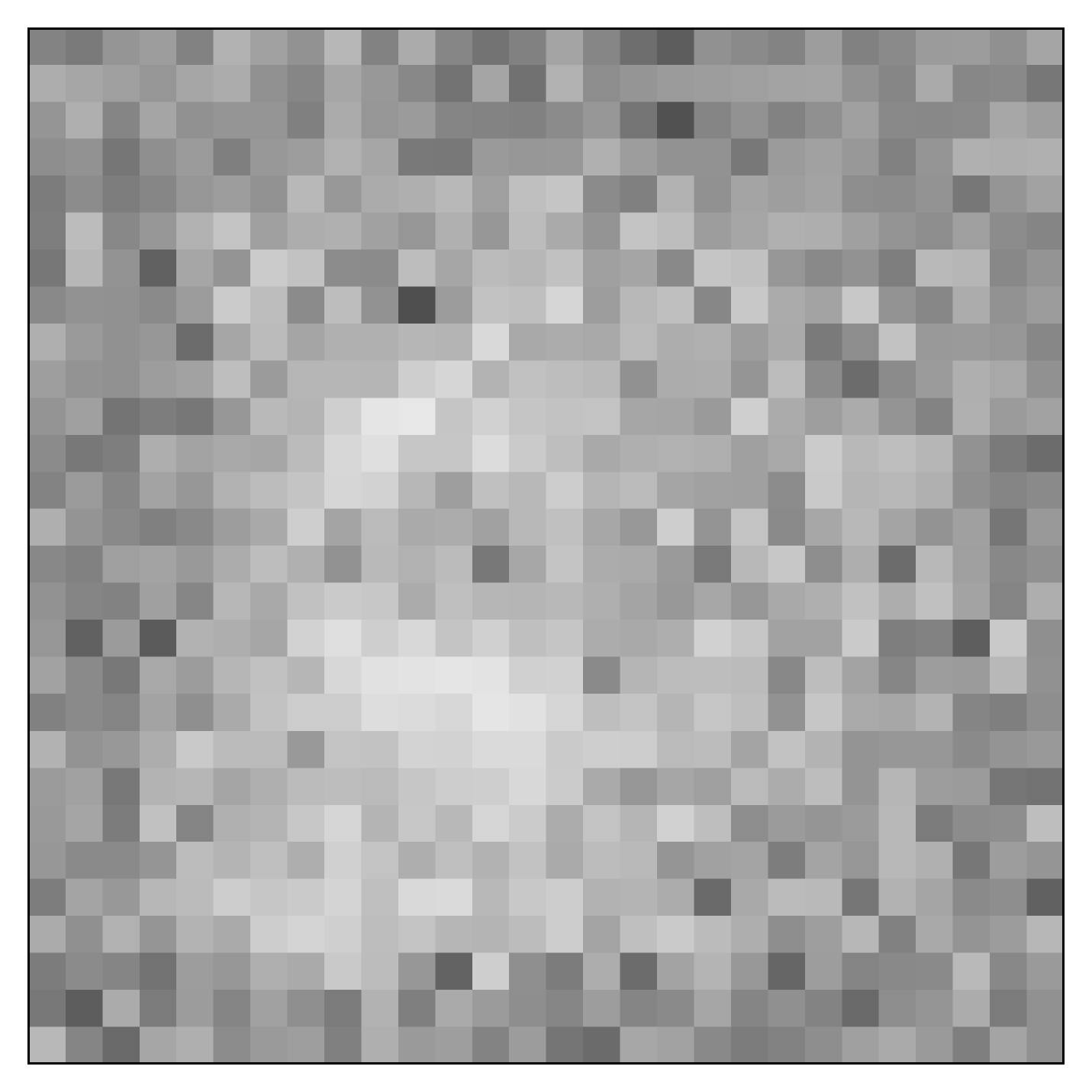}}
\subfigure{\includegraphics[scale = 0.18]{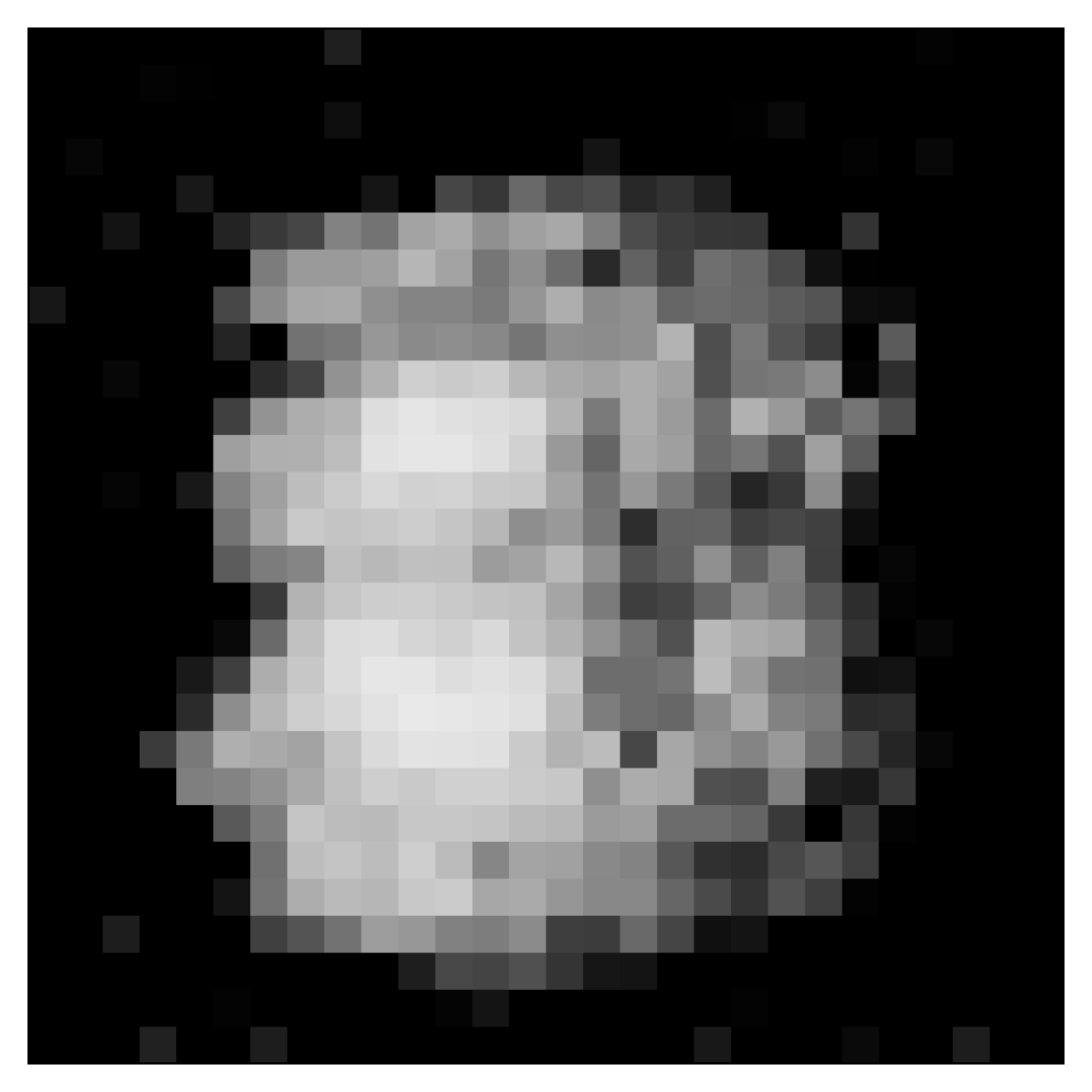}}
~~~~~
\subfigure{\includegraphics[scale = 0.18]{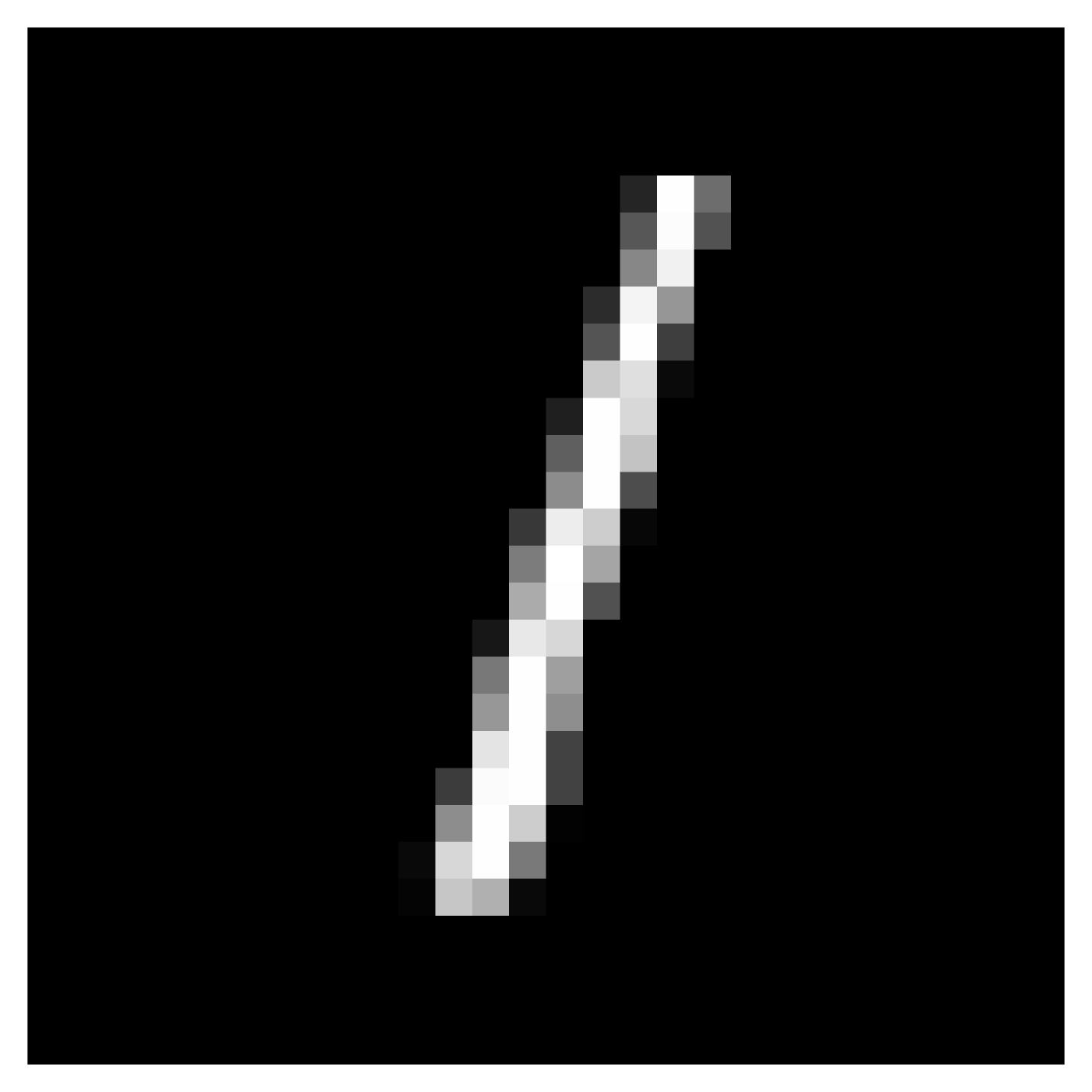}}
\subfigure{\includegraphics[scale = 0.18]{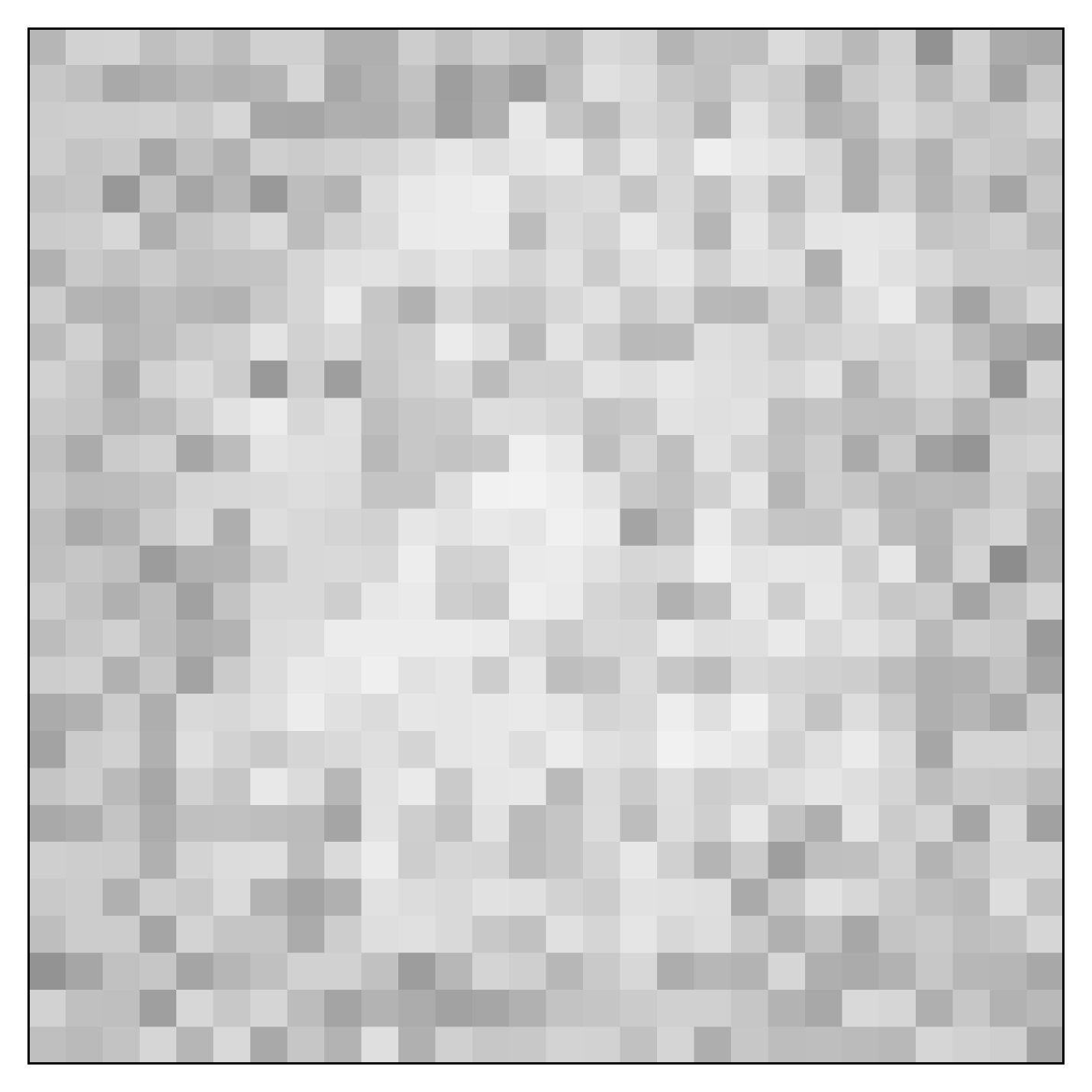}}
\subfigure{\includegraphics[scale = 0.18]{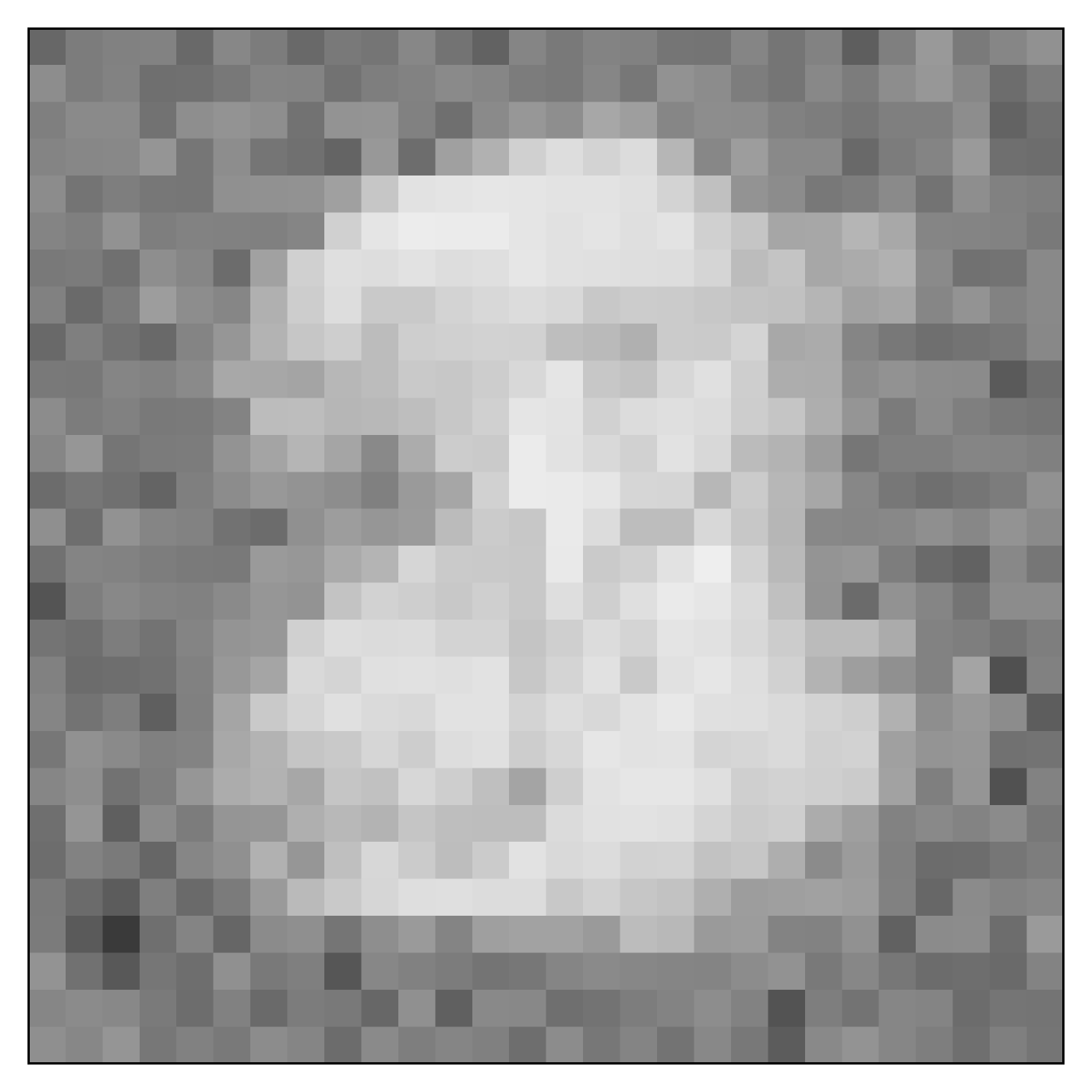}}
\\
\subfigure{\includegraphics[scale = 0.18]{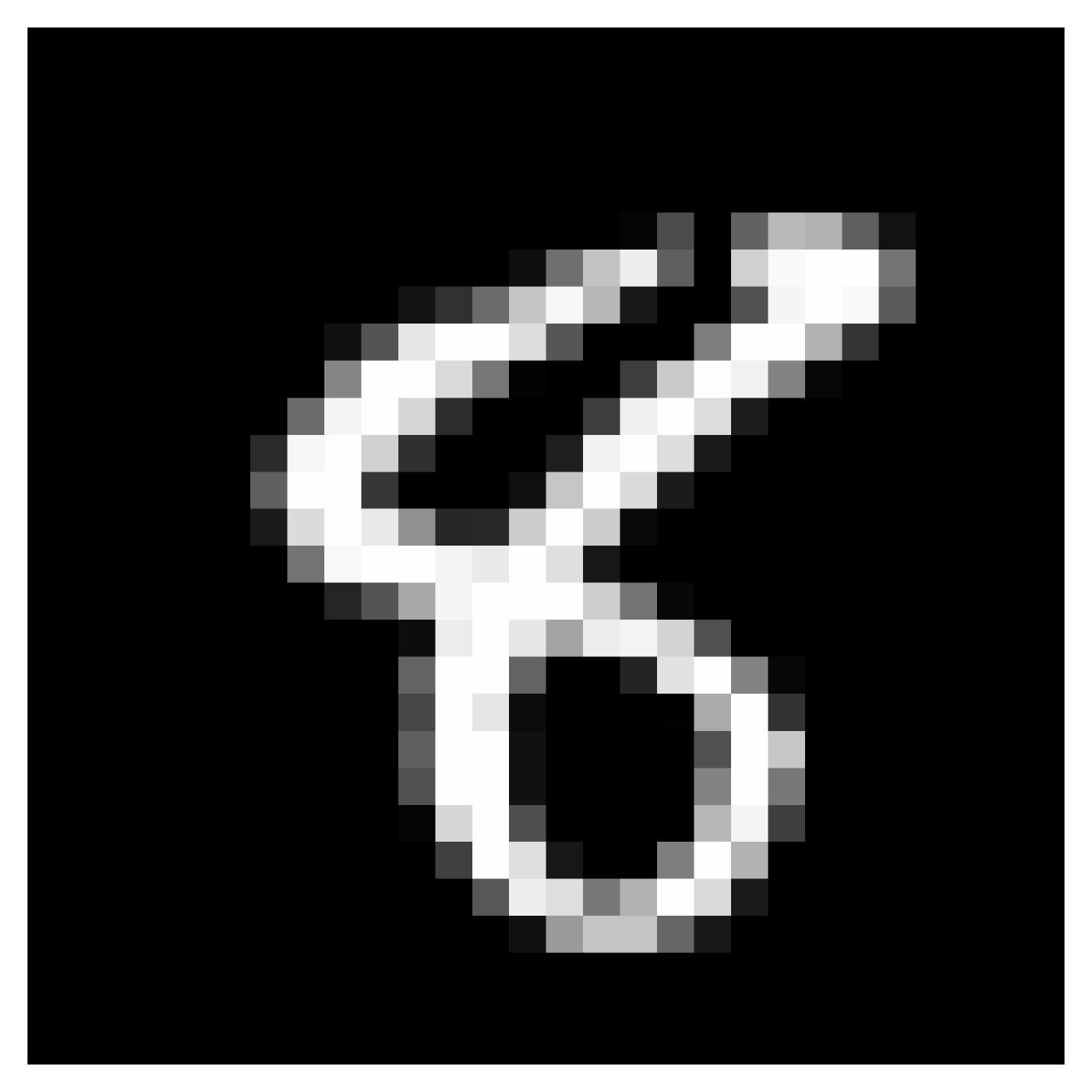}}
\subfigure{\includegraphics[scale = 0.18]{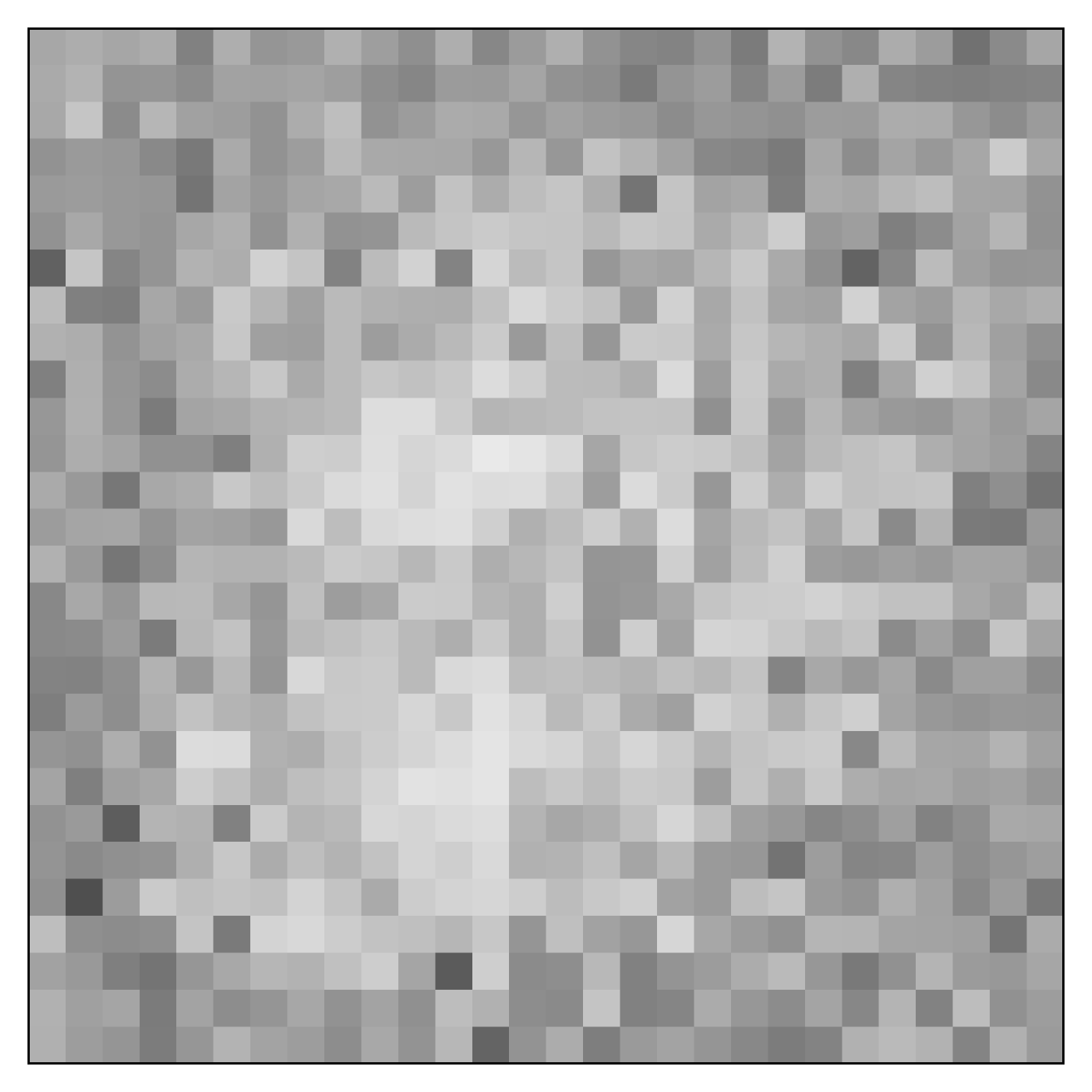}}
\subfigure{\includegraphics[scale = 0.18]{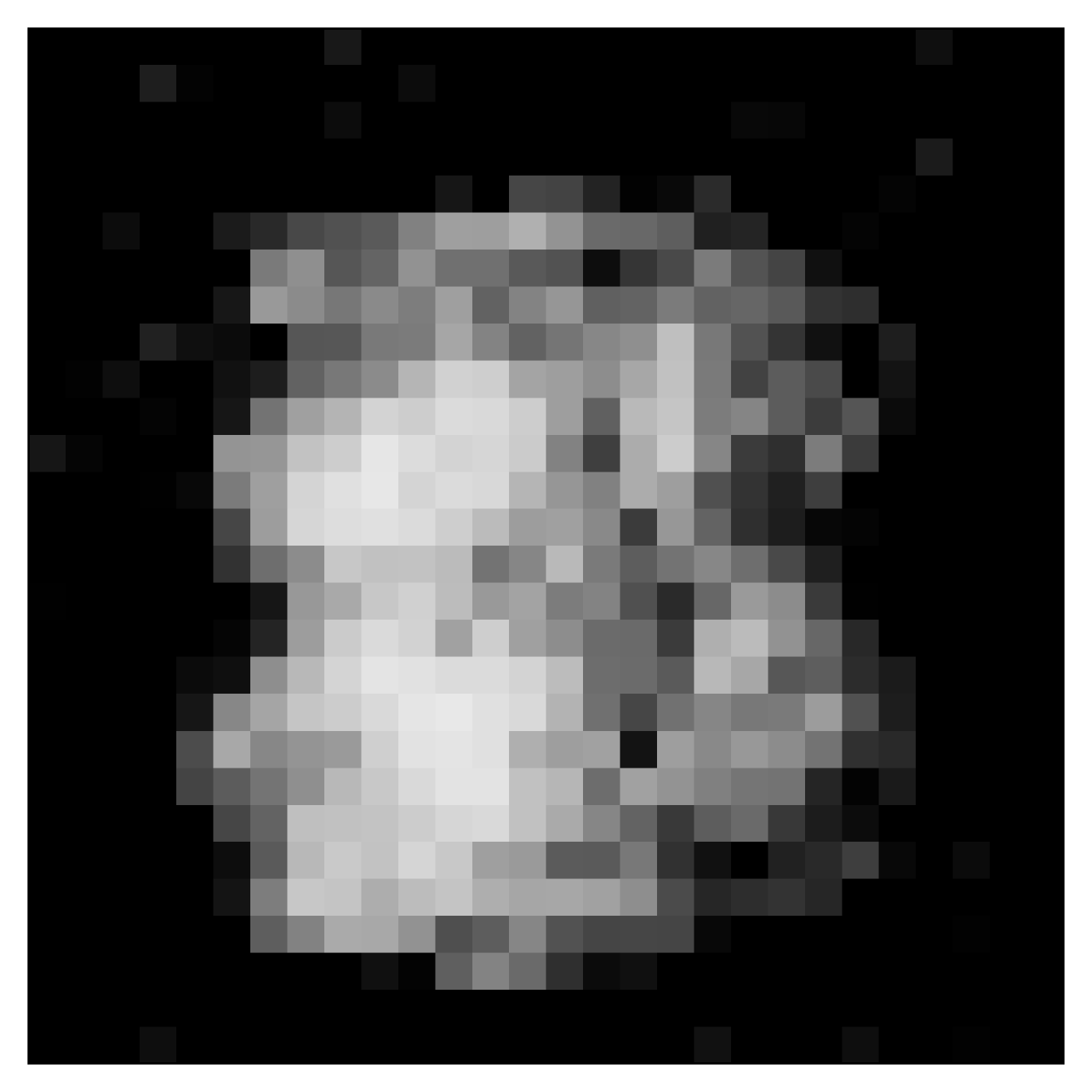}}
~~~~~
\subfigure{\includegraphics[scale = 0.18]{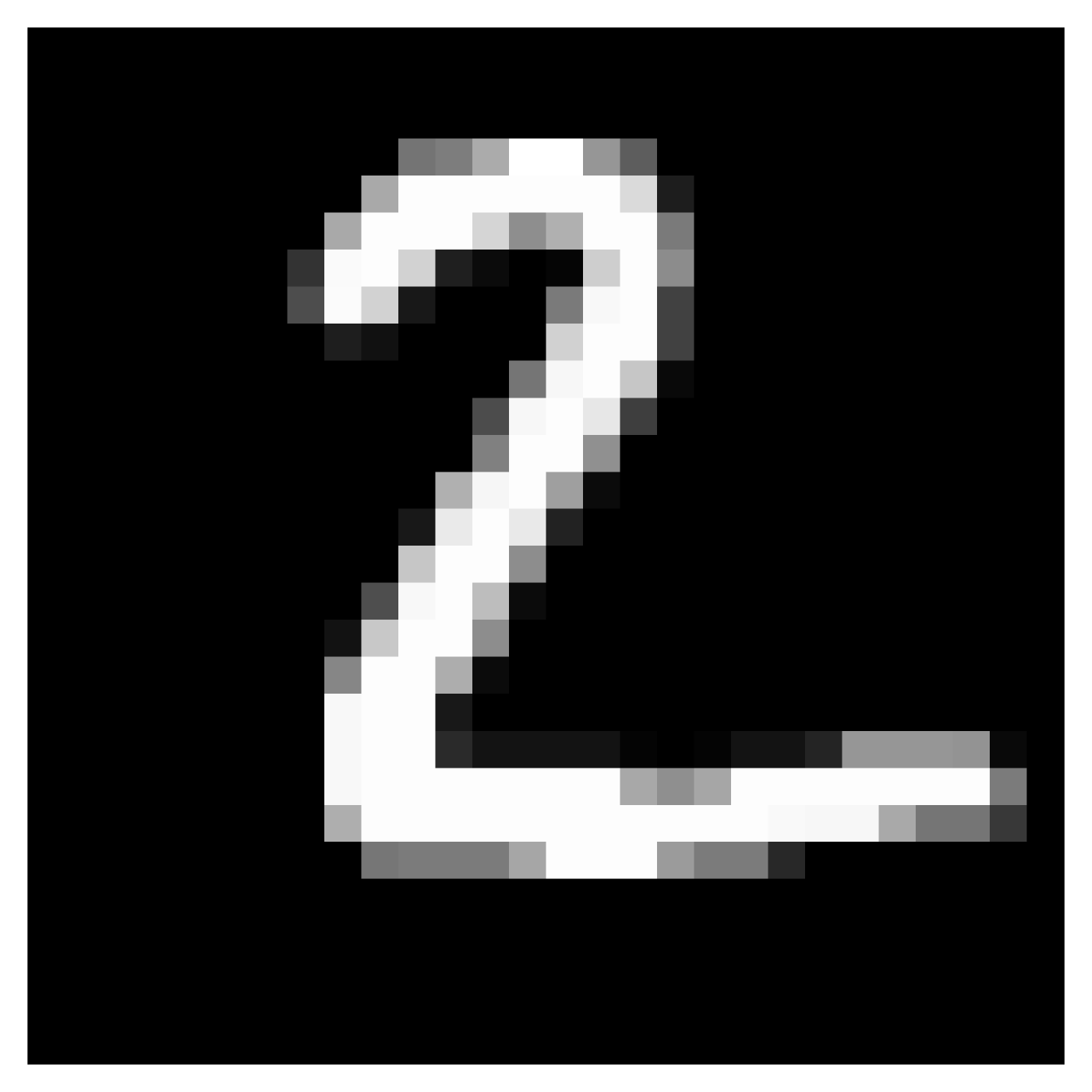}}
\subfigure{\includegraphics[scale = 0.18]{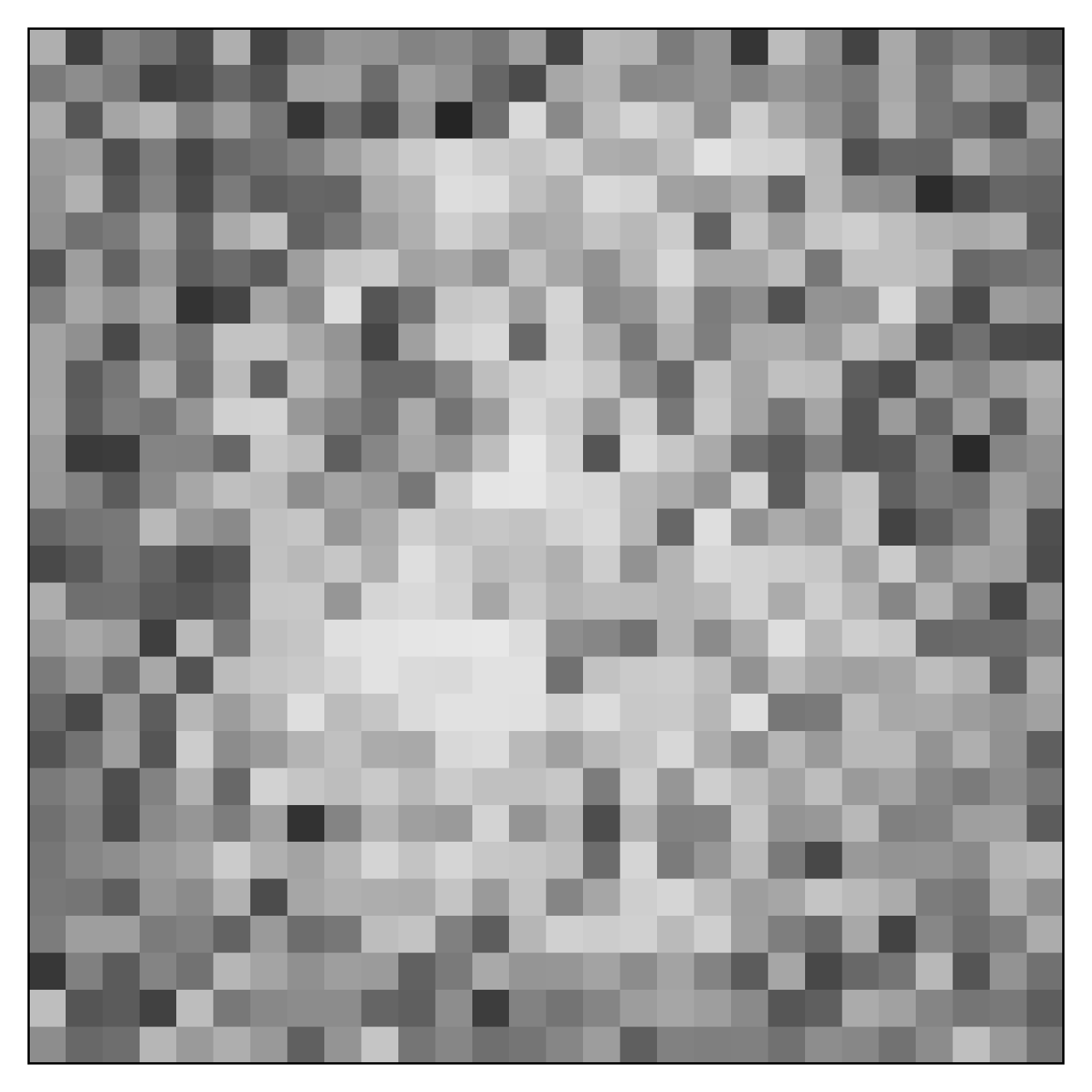}}
\subfigure{\includegraphics[scale = 0.18]{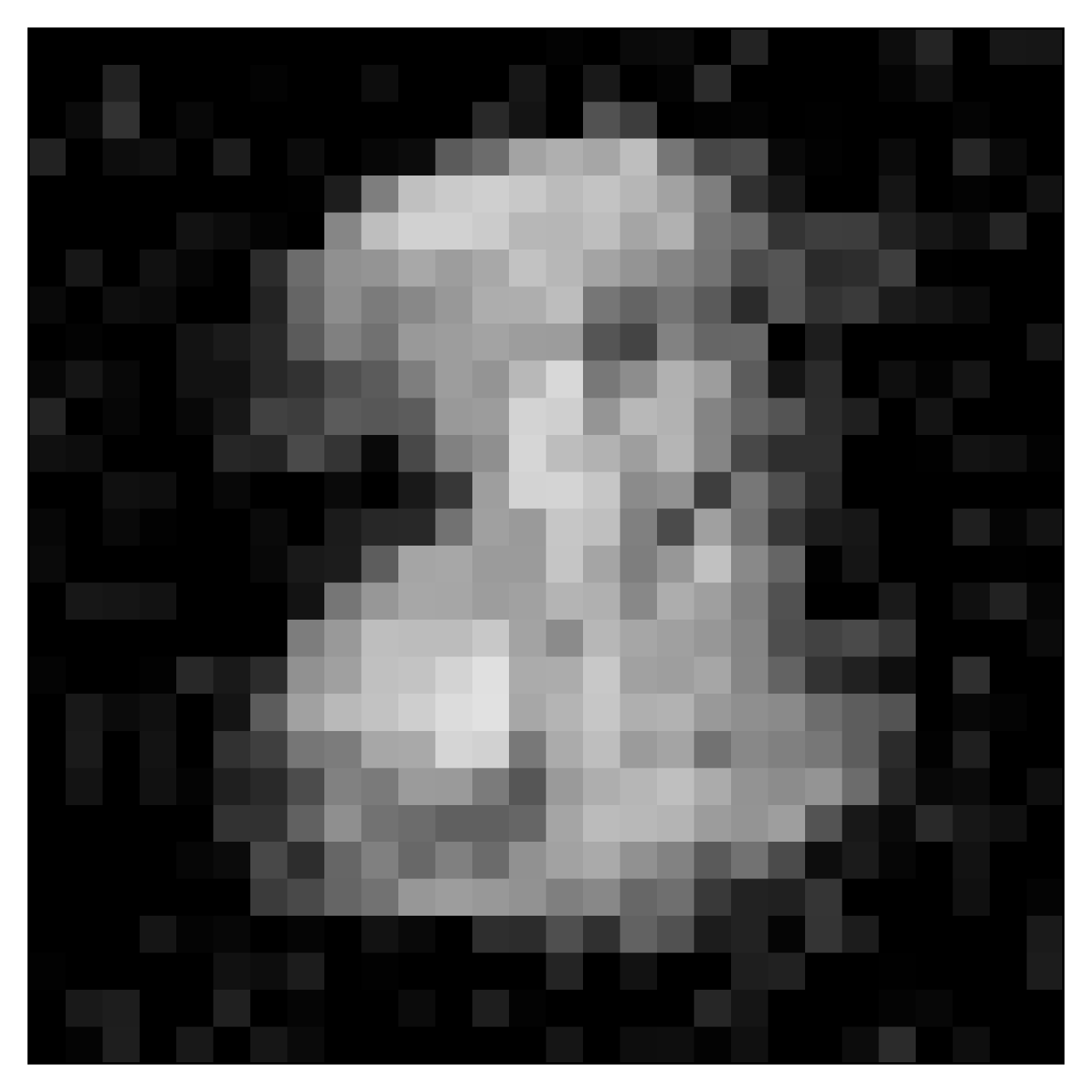}}
\caption{
Additional examples of bounding maps of images.
The images on the left are based on model distinguishing digit `3' from digit `8'.
The images on the right are based on model distinguishing digit `1' from digit `2'.
Each tuple of 3 images demonstrate the original images, the bounding map of the normal model and the bounding map of the robust model.
} \label{fig:app_bounding_maps}
\end{figure}

Figure \ref{fig:app_bounding_maps} gives more examples of bounding maps.
Like the model discussed in section \ref{sec:interpretability}, we train two model: one to distinguish digit `3' from digit `8', the other to distinguish digit `1' from digit `2'.

\subsection{Robustness and Decision Boundary} \label{sec:app_exp_sim}

We put the average and minimum values of cosine similarity between $\eps$ values of all image pairs in Table \ref{tbl:app_cos_similarity}.

\begin{table}[H]
\centering
\begin{tabular}{|c|c|c|c|c|}
\hline
Dataset & Architecture & Adversarial Training & Mean Cosine & Minimum Cosine \\
\hline
\multirow{6}{*}{MNIST} & \multirow{2}{*}{100-100-100} & -  & 0.9548 & 0.2304 \\
\cline{3-5}
& & PGD, $\tau = 0.1$ & 0.9957 & 0.9155 \\
\cline{2-5}
& \multirow{2}{*}{300-300-300} & - & 0.9774 & 0.5038 \\
\cline{3-5}
& & PGD, $\tau = 0.1$ & 0.9964 & 0.9104 \\
\cline{2-5}
& \multirow{2}{*}{500-500-500} & - & 0.9874 & 0.6367 \\
\cline{3-5}
& & PGD, $\tau = 0.1$ & 0.9941 & 0.8920 \\
\hline
\multirow{2}{*}{Fashion-MNIST} & \multirow{2}{*}{1024-1024-1024} & - & 0.9805 & 0.5652 \\
\cline{3-5}
& & PGD, $\tau = 0.1$ & 0.9752 & 0.7166 \\
\hline
\multirow{2}{*}{SVHN} & \multirow{2}{*}{1024-1024-1024} & - & 0.9836 & 0.7129 \\
\cline{3-5}
& & PGD, $\tau = 0.1$ & 0.9952 & 0.9339 \\
\hline
\end{tabular}
\caption{Cosine Similarity of $\eps$} \label{tbl:app_cos_similarity}
\end{table}

\section{Algorithm for General Architectures} \label{sec:app_architecture}

\subsection{General Linear Layers}

We first show our formulation for fully-connected layer in \textbf{Algorithm} \ref{alg:bound_est} can be naturally generalized to \textit{any linear layer} with fixed variables.
Here, we use popular convolutional layer as an example.

Given a convolutional layer with weights $\W^{(c)} \in \R^{n_{in} \times n_{out} \times f_w \times f_h}$, input $x_{in} \in \R^{n_{in} \times w_{in} \times h_{in}}$ and output $x_{out} \in \R^{n_{out} \times w_{out} \times h_{out}}$.
If we do not use any padding, then $w_{out} = w_{in} - f_w + 1$ and $h_{out} = h_{in} - f_h + 1$.
Therefore, we can reshape $x_{in}$, $x_{out}$ into one-dimensional vectors $x'_{in}$, $x'_{out}$ and rewrite convolutional operator in matrix multiplication form $x'_{out} = \W x'_{in}$.
$\W$ is a fixed matrix defined as follows where $i_1, i_2, j_1, j_2, k_1, k_2$ are integers.

\begin{equation}
\begin{aligned}
\W_{i_1 w_{in}h_{in} + j_1 h_{in} + k_1, i_2 w_{out}h_{out} + j_2 h_{in} + k_2} = \left\{
\begin{aligned}
&\W^{(c)}_{i_1, i_2, j_1 - j_2, k_1 - k_2}; 0 \leq j_1 - j_2 \leq f_w - 1, 0 \leq k_1 - k_2 \leq f_h - 1 \\
&0~~~~~~~~~~~~~~~~~~~~~~~~~~~~; \mathrm{otherwise}
\end{aligned}
\right.
\end{aligned}
\end{equation}

As a result, we will use matrix multiplication \footnote{We drop bias term for simplicity.} to represent \textit{any linear layer} defined in the neural networks.

\subsection{General Feedforward Neural Networks}

\begin{figure}[h!]
\centering
\includegraphics[scale = 0.5]{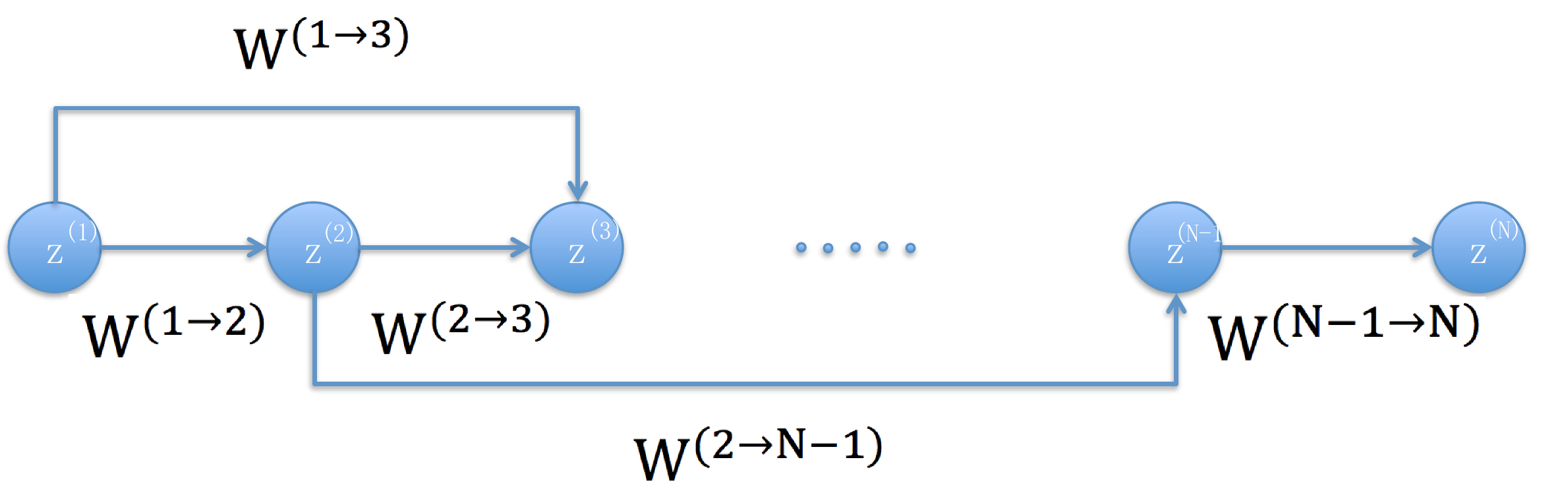}
\caption{An example of DAG representation of a general feedforward neural networks.} \label{fig:dag}
\end{figure}

As Figure \ref{fig:dag} shows, any general feedforward neural networks can be considered as directed acyclic graph (DAG) $G(V, E)$.
The vertices $\{1, 2, ..., N - 1, N\}$ represent neurons of each layer and the activation functions are applied on the internal nodes i.e. hidden units.
Any edge $(j, i) \in E$ corresponds to a weight matrix $\W^{(j \to i)}$ representing a direct connection between the output of layer $j$ and the input of layer $i$ ($i > j$).

Now we can define the formulation of a general feedforward neural network, which covers popular residual networks (ResNet) \cite{he2016deep} and densely connected networks (DenseNet) \cite{huang2017densely}.

\begin{equation}
\begin{aligned}
\z^{(i)} &= \sum_{j = 1}^{i - 1} \W^{(j \to i)} \hat{\z}^{(j)} + \bias^{(i)} \\
\hat{\z}^{(i)} &= \sigma(\z^{(i)})
\end{aligned}\label{eq:general_nn}
\end{equation}

Formulation (\ref{eq:general_nn}) can be reduced to (\ref{eq:nn}) if no skip-connection exists i.e. $\W^{(j \to i)} = 0\ \forall j \neq i - 1$.
For notation simplicity, we first define $\Path^{(n)}(j, i)$ as the set of all paths of length $n$ from vertex $j$ to $i$ ($i > j$) in graph $G(V, E)$:

\begin{equation}
\begin{aligned}
\Path^{(n)}(j, i) &= \{(p_0, p_1, ..., p_{n - 1}, p_n) | j = p_0 < p_1 < ... < p_{n - 1} < p_n = i, (p_t, p_{t + 1}) \in E\ \forall 0 \leq t \leq n - 1\} \\
\end{aligned}
\end{equation}

Then we can define the linearized composite transformation from layer $j$ to layer $i$ as follows:

\begin{equation}
\begin{aligned}
\M^{(j \to i)} &= \sum_{n = 1}^{i - j} \sum_{\path \in \Path^{(n)}(j, i)} \Pi_{t = 0}^{n - 1} \W^{(p_t \to p_{t + 1})}\D^{(p_t)}~~~~ \forall\ j < i
\end{aligned} \label{eq:general_composite}
\end{equation}

Specially, we define $\M^{(j \to i)} = \I\ \forall j = i$.
Similar to equation (\ref{eq:zi}), we can write the output of each layer $\z^{(i)}$ by:

\begin{equation}
\begin{aligned}
\z^{(i)} = \sum_{l = 2}^{i} \M^{(l \to i)}\W^{(1 \to l)}\x + \sum_{h = 1}^{i} \M^{(h \to i)}\bias^{(h)} + \sum_{h = 1}^{i - 1}\sum_{l = h + 1}^{i} \M^{(l \to i)}\W^{(h \to l)}\m^{(h)}
\end{aligned} \label{eq:general_zi}
\end{equation}

In equation (\ref{eq:general_composite}) and (\ref{eq:general_zi}), $\{\D^{(h)}, \m^{(h)}\}_{h = 1}^{N}$ are defined in the same way as linear approximation in Section \ref{subsection:nonlinear_approx}.

Based on equation (\ref{eq:general_zi}), we can design the bound estimation algorithm (\textbf{Algorithm} \ref{alg:general_bound_est}) for a general feedforward neural network.
In line \ref{line:general_m}, $\M^{(j \to i + 1)}$ can be calculated recursively as follows, so the total complexity is still quadratic in matrix multiplication.

\begin{equation}
\begin{aligned}
\M^{(j \to i + 1)} = \sum_{h = j + 1}^i \W^{(h \to i + 1)} \D^{(h)} \M^{(j \to h)}
\end{aligned}
\end{equation}

We can then follow the same path, gradient calculation and optimization by the augmented Lagragian method, to estimate the certified region of the largest volume for a given data point $\x$.

\begin{algorithm}
\small
\begin{algorithmic}[1]
\STATE Input: Parameters $\{\W^{(i)}, \bias^{(i)}\}_{i = 1}^{N - 1}$, perturbation set $\Set_{\eps}(\x)$.
\STATE $\phi^{(1)} = \x$, $\D^{(1)} = \I$, $\m^{(1)}_1 = -\eps$, $\m^{(1)}_2 = \eps$
\STATE $\low^{(2)} = \W^{(1)}\x - \W^{(1)}_{+}\eps + \W^{(1)}_{-}\eps + \bias^{(1)}$
\STATE $\up^{(2)} = \W^{(1)}\x - \W^{(1)}_{-}\eps + \W^{(1)}_{+}\eps + \bias^{(1)}$
\STATE $\M^{(1 \to 2)} = \W^{(1)}$
\STATE $\phi^{(2)} = \W^{(1)}\x + \bias^{(1)}$
\FOR {i = $2, ..., N - 1$}
    \STATE Calculate $\D^{(i)}$, $\m_1^{(i)}$, $\m_2^{(i)}$ based on $\low^{(i)}$ and $\up^{(i)}$.
    \STATE $\low^{(i + 1)}_{simp} = \sum_{j = 1}^{i} \left(\W^{(j \to i + 1)}_{+}\sigma(\low^{(j)}) + \W^{(j \to i + 1)}_{-}\sigma(\up^{(j)}) \right)$
    \STATE $\up^{(i + 1)}_{simp} = \sum_{j = 1}^{i} \left(\W^{(j \to i + 1)}_{-}\sigma(\low^{(j)}) + \W^{(j \to i + 1)}_{+}\sigma(\up^{(j)}) \right)$
    \STATE \label{line:general_m} Calculate $\M^{(j \to i + 1)}$ for $\forall j < i + 1$ according to equation (\ref{eq:general_composite}).
    \STATE $\phi^{(i + 1)} = \sum_{j = 1}^{i + 1} \W^{(j \to i + 1)}\D^{(j)}\phi^{(j)} + \bias^{(i + 1)}$
    \STATE $\low^{(i + 1)}_{quad} = \phi^{(i + 1)} + \sum_{h = 1}^{i - 1} \left[ \left(\sum_{l = h + 1}^{i} \M^{(l \to i)}\W^{(h \to l)}\right)_{+}\m^{(h)}_1 + \left(\sum_{l = h + 1}^{i} \M^{(l \to i)}\W^{(h \to l)}\right)_{-}\m^{(h)}_2 \right]$
    \STATE $\up^{(i + 1)}_{quad} = \phi^{(i + 1)} + \sum_{h = 1}^{i - 1} \left[ \left(\sum_{l = h + 1}^{i} \M^{(l \to i)}\W^{(h \to l)}\right)_{-}\m^{(h)}_1 + \left(\sum_{l = h + 1}^{i} \M^{(l \to i)}\W^{(h \to l)}\right)_{+}\m^{(h)}_2 \right]$
    \STATE $\low^{(i + 1)} = \max(\low^{(i + 1)}_{simp}, \low^{(i + 1)}_{quad})$
    \STATE $\up^{(i + 1)} = \min(\up^{(i + 1)}_{simp}, \up^{(i + 1)}_{quad})$
\ENDFOR
\STATE Output: Bounds $\{\low^{(i)}, \up^{(i)}\}_{i = 2}^{N}$
\end{algorithmic}
\caption{Bound Estimation for General Feedforward Neural Network} \label{alg:general_bound_est}
\end{algorithm}

\end{document}